\documentclass[times,twocolumn,final]{elsarticle}

\usepackage{medima}
\usepackage{framed,multirow}

\usepackage{amssymb}
\usepackage{latexsym}
\usepackage{amsmath}
\usepackage{multirow}
\usepackage{array}

\usepackage{url}
\usepackage{xcolor}
\usepackage{arydshln}
\usepackage{hyperref}

\journal{Medical Image Analysis}

\begin{document}

\verso{Sascha Jecklin \textit{et~al.}}

\begin{frontmatter}

\title{Domain adaptation strategies for 3D reconstruction of the lumbar spine using real fluoroscopy data}
\author[1]{Sascha \snm{Jecklin}\corref{cor1}}
\cortext[cor1]{Corresponding author}
\ead{sascha.jecklin(at)balgrist.ch}
\author[1]{Youyang \snm{Shen}}
\author[1]{Amandine \snm{Gout}}
\author[1]{Daniel \snm{Suter}}
\author[1]{Lilian \snm{Calvet}}
\author[1]{Lukas \snm{Zingg}}
\author[2]{Jennifer \snm{Straub}}
\author[1]{Nicola Alessandro \snm{Cavalcanti}}
\author[1]{Mazda \snm{Farshad}}
\author[1]{Philipp \snm{Fürnstahl}\corref{contrib}}
\author[1]{Hooman \snm{Esfandiari}\corref{contrib}}
\cortext[contrib]{These two authors contributed equally to this work}

\address[1]{Research in Orthopedic Computer Science, Balgrist University Hospital, University of Zurich, 8008 Zurich, Switzerland}
\address[2]{Universitätsklinik für Orthopädie, AKH Wien, Währinger Gürtel 18-20, 1090 Wien, Austrtia}

\received{}
\finalform{}
\accepted{}
\availableonline{}
\communicated{}

\begin{abstract}
In this study, we address critical barriers hindering the widespread adoption of surgical navigation in orthopedic surgeries due to limitations such as time constraints, cost implications, radiation concerns, and integration within the surgical workflow. Recently, our work X23D showed an approach for generating 3D anatomical models of the spine from only a few intraoperative fluoroscopic images. This approach negates the need for conventional registration-based surgical navigation by creating a direct intraoperative 3D reconstruction of the anatomy. Despite these strides, the practical application of X23D has been limited by a significant domain gap between synthetic training data and real intraoperative images.

In response, we devised a novel data collection protocol to assemble a paired dataset consisting of synthetic and real fluoroscopic images captured from identical perspectives. Leveraging this unique dataset, we refined our deep learning model through transfer learning, effectively bridging the domain gap between synthetic and real X-ray data. 
We introduce an innovative approach combining style transfer with the curated paired dataset. This method transforms real X-ray images into the synthetic domain, enabling the \emph{in-silico}-trained X23D model to achieve high accuracy in real-world settings.

Our results demonstrated that the refined model can rapidly generate accurate 3D reconstructions of the entire lumbar spine from as few as three intraoperative fluoroscopic shots. The enhanced model reached a sufficient accuracy, achieving an $84\%$ F1 score, equating to the benchmark set solely by synthetic data in previous research. Moreover, with an impressive computational time of just 81.1 ms, our approach offers real-time capabilities, vital for successful integration into active surgical procedures.

By investigating optimal imaging setups and view angle dependencies, we have further validated the practicality and reliability of our system in a clinical environment. Our research represents a promising advancement in intraoperative 3D reconstruction. This innovation has the potential to enhance intraoperative surgical planning, navigation, and surgical robotics.
\end{abstract}

\begin{keyword}
\MSC 41A05\sep 41A10\sep 65D05\sep 65D17
\KWD 3D reconstruction\sep intraoperative \sep Deep Learning\sep X-ray\sep Fluoroscopy
\end{keyword}

\end{frontmatter}

\section{Introduction}
\label{sec:intro}
Intraoperative navigation of complex orthopedic interventions remains challenging despite modern computer-assisted surgery (CAS) solutions. The adoption rate of CAS solutions in the orthopedic surgical practice has been reported to be alarmingly low, estimated to be approximately in the range of 11\% \cite{hartl_worldwide_2013} or 5\% \cite{joskowicz_computer_2016} of surgeries performed. 
This is particularly concerning considering the large body of documented evidence that demonstrates the benefits of CAS solutions in improving surgical accuracy and outcomes \cite{kelley_utilization_2021, cui_application_2012, gelalis_accuracy_2011, mirza_accuracy_2003, tian_pedicle_2011, wang_computer-assisted_2019}.
In the context of implant positioning for spine surgery, a recent meta-analysis investigating the accuracy of 37'337 pedicle screw placements \cite{perdomo-pantoja_accuracy_2019} highlights the impact of surgical navigation. Further analysis of the results reported in \cite{perdomo-pantoja_accuracy_2019,esfandiari_intraoperative_2020} reveals that the per implant malplacement rate decreases from 10.2\% in cases without surgical navigation to 4.9\% when computed tomography (CT) navigation is used.
Despite these advantages, the limited adoption of CAS solutions in clinical practice can be attributed to additional factors. These include increased surgical time \cite{wang_review_2020}, operational costs \cite{dea_economic_2016}, line of sight issues (specific to solutions that require optical tracking systems) \cite{mehbodniya_frequency_2019}, and most importantly, the need for patient registration \cite{chiang_computed_2012, nakanishi_usefulness_2009}.

Our approach provides an alternative to existing CAS solutions by focusing on improving intraoperative imaging capabilities. Unlike traditional CAS methods that rely on preoperative data and complex patient registration processes, our method leverages real-time intraoperative data to create accurate 3D reconstructions of the spine. This approach can be seamlessly integrated with current intraoperative imaging systems, enhancing their functionality without the need for extensive preoperative preparation.

Patient registration, which is at the heart of a common CAS solution, involves aligning a preoperatively generated volumetric 3D representation of the anatomy (e.g., a CT scan) to the intraoperative position of the patient and has been addressed in previous research using various algorithms. These include feature-based registration \cite{groher_segmentation-driven_2007, xin_chen_extension_2006}, intensity-based registration \cite{penney_validation_2001, esfandiari_comparative_2019, tang_hardware-assisted_2004}, RGB-only registration \cite{liebmann_automatic_2024}, statistical shape modeling (SSM) \cite{pavlova_statistical_2017, baka_2d3d_2011}, and, more recently, data-driven deep learning registration methods \cite{miao_cnn_2016, miao_dilated_2018, liao_artificial_2017, ferrante_adaptability_2018, zheng_pairwise_2018}.

Despite the availability of advanced algorithms for patient registration, commercially available CAS solutions often initially rely on rudimentary landmark-based methods. This is mainly due to their ease of integration into CAS pipelines compared to more advanced registration methods, which suffer from limiting factors such as a small capture range \cite{esfandiari_comparative_2019}, limited training datasets \cite{haskins_deep_2020}, inadequate similarity metrics for multimodality registration \cite{haskins_learning_2019}, and long computation times \cite{mcdonald_comparison_2007}.

There is a growing interest in developing registration-free computer-assisted surgery (CAS) alternatives, which aim to create an up-to-date 3D representation of the patient's anatomy during the intervention, utilizing only the available intraoperative data. Such methods can practically reduce the preoperative preparation stages, which can be seen as a shortcoming in existing CAS solutions. Furthermore, given their intraoperative nature, these emerging algorithms will not require the patient registration step, as all the spatial data required for surgical navigation is created intraoperatively. This is particularly relevant in urgent cases where no preoperative data is available (e.g., an orthopedic trauma surgery). In the context of orthopedic surgery, registration-free CAS can be achieved using two primary approaches. First, modern C-arm devices can be used that are equipped with cone beam computed tomography (CBCT) features to achieve an up-to-date 3D representation of the underlying anatomy \cite{tian_pedicle_2011} by a direct 3D volumetric reconstruction. The adoption rate of CBCT-based CAS solutions has been limited due to factors such as increased cost \cite{dea_economic_2016,tonetti_role_2020}, extended procedure time \cite{beck_benefit_2009}, and ionizing radiation exposure \cite{costa_spinal_2011}.

As an alternative to CBCT-based CAS, recent advances in machine learning have materialized the possibility of achieving intraoperative 3D representations based on sparse intraoperative data. For example, recent developments in the field include methods such as \cite{fang_3d_2020}, which can reconstruct the patient's spine based on a few available intraoperative fluoroscopy shots. Similarly, work reported in \cite{kasten_end--end_2020, shiode_2d3d_2021} allow for reconstructions of the knee and wrist, respectively. We recently published X23D, our inaugural article on intraoperative 3D representation of spine anatomy based on sparse fluoro data \cite{jecklin_x23dintraoperative_2022}. 
Our study offers an innovative approach to enhancing intraoperative imaging systems by providing accurate 3D reconstructions of the spine based on real-time sparse fluoroscopy data. This method complements existing surgical navigation systems, such as Ziehm NaviPort (Nuremberg, Germany) or eCential Robotics (Gières, France), by enhancing their imaging capabilities and potentially integrating seamlessly into their workflows.
The results have shown promising performance, highlighting its potential utility in the field. However, the limited availability of high-quality, large-scale, annotated intraoperative data presents a significant technical bottleneck in further developing such methods. Ethical concerns, radiation exposure, and the manual effort required for data annotation are obstacles to collecting the necessary \textit{in-vivo} training datasets for these algorithms. For these reasons, our prior work entirely relied on \textit{in-silico} intraoperative fluoroscopy datasets.
These \textit{in-silico} datasets can be created using digitally reconstructive radiograph (DRR) techniques \cite{gertzbein_accuracy_1990, unberath_deepdrr_2018, li_digitally_2006} with varying levels of fidelity. DRR techniques simulate the projection of a planar radiograph based on an input CT volume, offering flexibility and scalability in dataset generation. 
The use of synthetic data circumvents the lack of labeled data but comes with a domain gap between the \textit{in-silico} training domain and the \textit{in-vivo} target domain. This limitation highlights the motivation and necessity of the present study to bridge the gap between an \textit{in-silico} trained intraoperative deep learning model and its use in real surgeries.

The domain gap phenomenon stems from the fact that DRRs cannot capture realistic intraoperative imaging conditions, such as the anatomy's true radiological properties, the imaging device's noise characteristics, and the imaging environment's backscattering behavior. Consequently, when these \textit{in-silico} datasets are used to train downstream deep learning models, the models tend to perform well only on the \textit{in-silico} data but struggle to achieve similar performance levels when applied to real intraoperative X-ray images. 

Recent advancements, such as DeepDRR \cite{unberath_deepdrr_2018}, have attempted to bridge this domain gap by incorporating realistic X-ray characteristics like attenuation, scatter, and noise into the DRR generation process. DeepDRR has shown significant improvements in realism, making it a valuable tool in specific intraoperative settings, as demonstrated by \cite{kausch_toward_2020}. However, the added complexity of simulating detailed X-ray physics can introduce new challenges, particularly in handling the heterogeneity of data arising from various C-arms and clinical environments.

Our approach differs in how it handles data heterogeneity, which is inevitable when intraoperative data is targeted. Instead of simulating complex X-ray physics, we focus on extracting essential bone structures to create straightforward intermediate representations. This method reduces the impact of data heterogeneity, providing a more generalized solution that is robust across different imaging conditions. 
Our DRR efforts produce images in a straightforward manner that emphasizes essential anatomical features without the added complexity of simulating noise and scatter. This strategy simplifies the downstream training process and enhances the model's generalizability to different imaging devices and settings.

In the broader computer vision community, there are two predominant families of techniques to handle the domain gap phenomenon: transfer learning \cite{srivastav_improved_2021} and style transfer \cite{kausar_sd-gan_2023}. In our work we evaluate both of these techniques separately to connect our intermediate representations to real-world X-rays. By using transfer learning, we adapt synthetically pre-trained models to our specific task with minimal additional training on real data. Concurrently, style transfer techniques enable the transformation of real X-rays to match the intermediate representations generated by our approach, thereby bridging the gap between synthetic and real data in a practical and effective manner.

Style transfer has been a part of an active field of research with the advent of novel deep learning-based methods \cite{jing_neural_2020,deng_stytr_2022} that have recently outperformed traditional approaches. These methods can generally be categorized based on the type of training data they require: paired \cite{yang_learning_2020, xu_semi-paired_2021} or unpaired data \cite{zhang_domain_2022,yi_dualgan_2017}. 
Style transfer can be used to increase the fidelity level of pre-existing (i.e., \textit{in-silico}) data, improving the generalizability of the downstream deep learning models. This is achieved by combining the semantic content of a source image (i.e., a DRR image) with the texture information of a target image (i.e., an X-ray) \cite{gatys_image_2016}. Our specific goal in this study is to investigate the feasibility of the mentioned domain adaptation techniques for our target application of intraoperative AI-based 3D reconstruction.   

Studies such as \cite{alzubaidi_novel_2021} and \cite{kim_artificial_2018} have highlighted the advantages of employing transfer-learned networks in the medical domain, even when the networks were not initially trained on medical data. For instance, \cite{abdel-aziz_direct_2015} demonstrated the effectiveness of this approach in skin cancer classification, while \cite{kim_artificial_2018} reported successful results in fracture detection.
In the realm of intraoperative X-ray data analysis, transfer learning strategies have focused on harnessing knowledge from synthetic datasets to improve performance on real X-ray images. Notable examples include the use of GANs as demonstrated in \cite{srivastav_improved_2021} for pneumonia detection, and the application of DRRs as explored in \cite{wang_multi-view_2021} for registration tasks. Both approaches first utilized \textit{in-silico} generated data before refining the network and closing the domain gap by transfer learning on target domain data. 

Style transfer techniques developed for intraoperative use cases have focused on transferring images from the \textit{in-silico} domain, such as DRRs, to the X-ray domain \cite{gao_synthex_2022}. \cite{gao_synthex_2022} showcased the benefits of realistic simulation of image formations from human models on three examples: hip imaging, tool detection, and COVID lesion segmentation.
Even under controlled imaging conditions, X-ray images vary greatly in image and radiological properties. This depends on factors such as imaging technology, detector type (intensifier vs. flat panel) \cite{sheth_mobile_2018}, device settings \cite{tsalafoutas_evaluation_2021}, patient's Body Mass Index (BMI) \cite{cushman_effect_2016}, imaging view-angle, and back-scattering. This makes the task of DRR to X-ray transfer inherently challenging, given that the target domain in this setup (i.e., the X-ray domain) is highly heterogeneous.

A key contribution of our study to address this limitation is to use style transfer in the reverse direction, transferring X-ray images into the \textit{in-silico} domain. This unconventional approach is motivated by several key considerations. 
Our method creates DRR images with a homogeneous target domain, transforming data from a highly detailed to a lowly detailed domain. Despite improvements in DRR technology, DRRs are still an approximation of X-rays. The style transfer approach cannot account for the presence of anatomical features inaccurately portrayed in DRRs and would need to generate such features from the training data distribution. More importantly, this approach enables the application of both existing and future \textit{in-silico} trained models on transferred X-rays, allowing for more focused and effective optimization of each component.

A particular challenge for style transfer methods is collecting and curating paired data for training. This has prompted the introduction of more advanced methods capable of training even with unpaired data in recent years. Although these models offer greater flexibility in terms of their training data requirements, they generally underperform similar methods trained solely on paired data \cite{tripathy_learning_2018, xu_semi-paired_2021}. 
\cite{shiode_2d3d_2021} demonstrated an approach using style transfer capable of producing 3D reconstructions of the forearm from a single X-ray. To generate a substantial amount of synthetic data paired with X-rays, they incorporated a DRR process that used intensity-based registration. The DRRs were additionally altered to remove areas not part of the reconstruction target, focusing specifically on the ulna. While effective for their specific task, this method relies on precise registration, which can be computationally intensive and prone to errors. These potential registration errors could affect the accuracy of the final reconstruction. Moreover, by integrating the segmentation operation directly into the style transfer process, the method of \cite{shiode_2d3d_2021} becomes a tool for specific reconstructions but lacks versatility for broader applications.

As a significant contribution of this study, we present our novel data collection pipeline, which allows the creation of a paired DRR-X-ray dataset from several \textit{ex-vivo} human specimens. 
The core of our methodological contribution lies in the development of a custom protocol for the collection of paired DRR/X-ray datasets, which is a crucial enabler for effective and accurate style transfer. This unique dataset not only facilitates the use of the Pix2Pix algorithm in our context but also addresses a significant gap in current research by providing a means to directly improve the practical application of deep learning models in clinical settings. Furthermore, we argue that the integration of a robust data collection methodology with an established algorithm like Pix2Pix allows for a transformative improvement in the model's transition to clinical use. This aspect of our work represents a significant stride toward the real-world application of deep learning technologies in medicine.
Our work leverages the generated paired dataset and the homogeneity of the target domain to adopt the Pix2Pix model \cite{isola_image--image_2018}, a well-known conditional generative adversarial network (cGAN) architecture tailored for image-to-image translation tasks. We opted for the Pix2Pix model to translate X-ray images into the DRR domain, emphasizing its ability to preserve crucial features, including edges and content, while introducing stylistic changes that align with the target domain.

We have integrated the transfer learning and style transfer approaches into our existing 3D reconstruction pipeline, allowing for the application of X23D to real fluoroscopy data. To the best of our knowledge, this represents the first instance of successfully transitioning spine DRR-trained networks to real data.

The specific contributions of this study are: 

\begin{enumerate}
    \item \textbf{Dataset Acquisition Protocol:} This work outlines a protocol for collecting and processing clinical data, focusing on creating a dataset of paired real and synthetic X-ray images from \textit{ex-vivo} human specimens, crucial for our intraoperative domain adaptation.
    \item \textbf{Domain Adaptation:} The study presents domain adaptation methods for deep learning models trained on synthetic intraoperative data, emphasizing style transfer and transfer learning. These techniques effectively narrow the gap between synthetic and real X-ray data, boosting pre-trained models' real-world applicability and generalizability.
    \item \textbf{Experimental Analysis on Real X-ray Data:} The study conducts a comprehensive investigation of the performance of the X23D reconstruction model proposed in \cite{jecklin_x23dintraoperative_2022} when applied to real X-ray data. By evaluating the efficacy and accuracy of the method in a real-world context, this analysis provides valuable insights into the practical utility and limitations of the approach.
\end{enumerate}

\section{Methods}
\label{sec:methods}
The remainder of this paper is structured as follows. First, we present our \textit{in-silico} training data generation process in Section \ref{sec:method:synthetic_data}. This is followed by our clinical data collection and processing protocol in Section \ref{sec:method:real_data}, which enables the creation of paired synthetic-real X-ray images and the corresponding imaging parameters (intrinsic and extrinsic).
A localization process and the architecture of our 3D reconstruction model are introduced in Section \ref{sec:method:localization} and \ref{sec:method:network}.
Section \ref{sec:methods:insilico} introduces the \textit{in-silico} training with Section \ref{sec:methods:domad} focusing on closing the domain gap between our \textit{in-silico} trained model and real data. This section explores a style-transfer approach and transfer learning of our existing 3D reconstruction model X23D.
We present our performance evaluation in Section \ref{sec:methods:performance_eval}. First, we outline the data flow through our style-transfer, localization, and 3D reconstruction pipelines. We then investigate the sensitivity of our network to varying input configurations and assess our capability to close the domain gap on real data. Performance metrics and experimental details are elaborated upon in this section. 
For a comprehensive understanding of our work, the implementation framework is described in detail in Section \ref{sec:methods:implementation}. This section provides information on the technical aspects, training, and tools used to develop and implement our proposed methods.

\subsection{Synthetic Data Generation}
\label{sec:method:synthetic_data}
\begin{figure*}
    \centering
    \includegraphics[width=\textwidth]{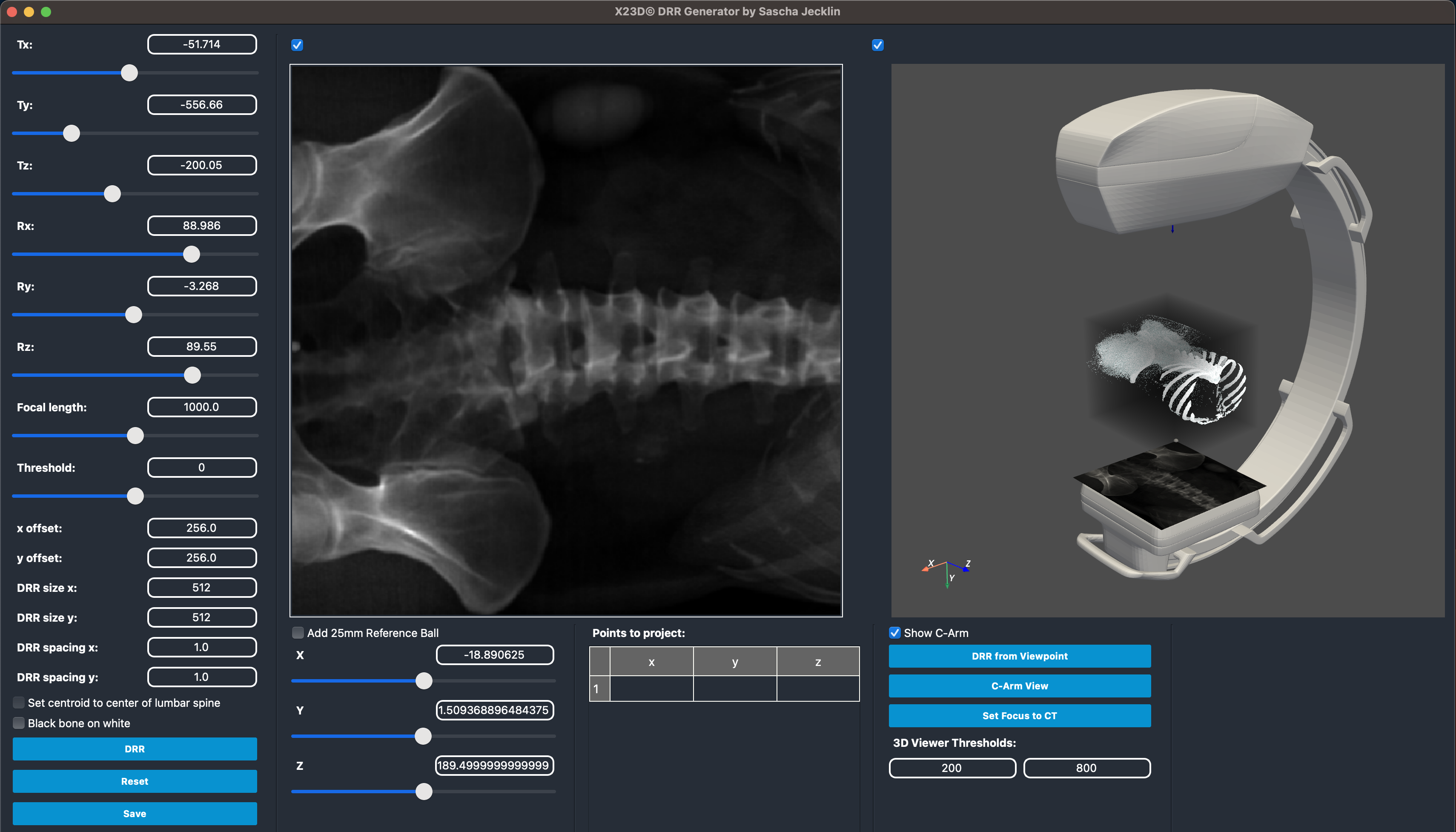}
    \caption{Frontend of our custom-built DRR generator software. Intrinsic and extrinsic parameters can be adapted on the left-hand side. The resulting DRR and the pose of the virtual C-arm are displayed next to it.}
    \label{fig:gui}
\end{figure*}
The front end of our custom-built data generation software is depicted in Figure \ref{fig:gui}. DRRs can be generated from arbitrary viewpoints around a CT scan. Intrinsic and extrinsic parameters can be adapted to simulate different characteristics of the C-arm, such as focal length. The DRR image $I$ is obtained by the following equation:

\begin{equation}
    I(p) = \int A(\mathsf{T}^{-1} L(p, s)) ds.
    \label{eq:drr}
\end{equation}

where $I(p)$ represents the intensity of the DRR at image point $p$. $\mathsf{T} = \left[
\begin{array}{c;{2pt/2pt}c}
    \mathsf{R} & \mathbf{X}_o 
\end{array}
\right] : \mathbb{R}^3 \rightarrow \mathbb{R}^3
$ is the transformation between the object and the image plane, $\mathbf{X}_o$ is the position of the world origin in camera coordinates, $\mathsf{R}$ is the rotation matrix whose columns are the direction of the world axes in the camera reference frame, $L(p, s)$ is the ray originating from the source of the X-rays and intersecting with the image plane at $p$, parameterized by $s$.

To ensure that only relevant anatomical structures are projected onto the DRR images, we specifically isolate bone structures by applying a thresholding technique to the underlying CT images. Concretely, we used a fixed Hounsfield value threshold of 0.

To generate synthetic DRRs, we selectively sampled 28 clinically feasible view angles per lumbar vertebrae of the CTs by capturing viewpoints from a sphere with a 1 meter diameter. The focal length for most C-arms typically falls close to 1 meter. Hence, the sphere with a $0.5$ m radius places the CT roughly in its isocenter. We varied the focal length during our \textit{in-silico} data generation to mimic different C-arm devices. During the DRR generation, four different classes of view angles were created to cover different X-ray perspectives:

\begin{itemize}
    \item \textbf{Anterior-posterior\,(AP)}: Contains both anterior-posterior and posterior-anterior images. Six images were sampled in the sagittal plane, with three images per side, deviating $-15^{\circ}$, $0^{\circ}$, and $+15^{\circ}$.
    \item \textbf{Lateral}: Includes images from both left and right lateral views. Six images were sampled in the coronal plane, with three images per side, deviating $-15^{\circ}$, $0^{\circ}$, and $+15^{\circ}$.
    \item \textbf{Oblique}: Comprises images that deviate in the transverse plane. Four images at $\pm20^{\circ}$ on each side of the CT.
    \item \textbf{Miscellaneous}: Contains images sampled outside the anatomical planes. We created 12 clusters of poses from the real dataset introduced in Section \ref{sec:method:real_data}. Each cluster's center corresponds to one miscellaneous pose in the synthetic dataset.
\end{itemize}

The CTSpine1K dataset \cite{deng_ctspine1k_2021} served as the base for our \textit{in-silico} dataset, providing us with 1005 CT scans of patients in DICOM format along with their vertebral level segmentations. This dataset contained 4712 lumbar vertebrae. We applied rigorous filtering based on specific inclusion and exclusion criteria to ensure data quality and relevance.
Our inclusion criteria encompassed CT scans with a resolution better than 1 mm in all three axes and CT volumes larger than 128 voxels on each axis. We excluded cases with pathological bone fusion (synostosis). After implementing the filtering process, the data set was refined to include 863 individual patient CT scans. These scans collectively accounted for 4291 vertebrae, forming the foundation of the synthetic dataset.
Centered on each of the 4291 vertebrae, we used the 28 poses mentioned above to generate DRRs. 

The resulting dataset of 120'148 DRRs and poses was split in the following way:
\begin{itemize}
    \item \textbf{Training}: 604 CT scans, 3000 vertebras, 84000 images
    \item \textbf{Validation}: 173 CT scans, 862 vertebras, 24136 images
    \item \textbf{Testing}: 86 CT scans, 429 vertebras, 12012 images
\end{itemize}

\subsection{Real Data Collection and Calibration}
\label{sec:method:real_data}
\begin{figure*}
    \centering
    \includegraphics[width=\textwidth]{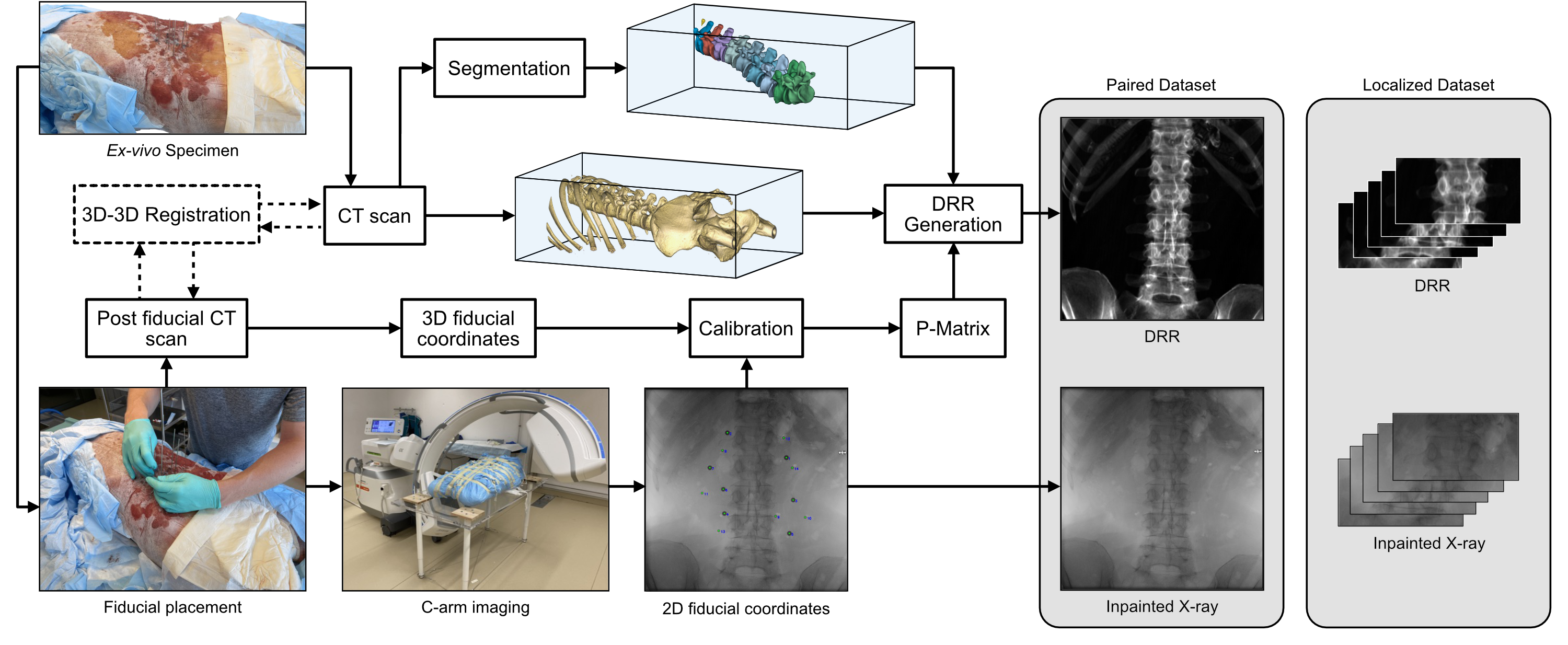}
    \caption{Our data collection process generates synthetic-real X-ray image pairs with pose information.}
    \label{fig:data_pipeline}
\end{figure*}

In addition to the generated \textit{in-silico} dataset, we constructed a data collection pipeline to acquire a paired \textit{ex-vivo} dataset containing real and synthetic X-ray images (collected from the same view angle) and their corresponding pose information (Figure \ref{fig:data_pipeline}).

The data collection process began by acquiring a high-resolution CT scan (SOMATOM Edge Plus, Siemens Healthcare, Erlangen, Germany, slice thickness: 0.75 mm, in-plane resolution: 0.5 mm × 0.5 mm) of a fresh-frozen entire torso \textit{ex-vivo} specimen. Vertebrae segmentation was then performed on the collected CT scan by trained specialists using Mimics software (Version 25.0, Materialise NV, Leuven, Belgium).

The age of the specimens ranged from 55 to 89 years, with a median age of 75 years. The BMI values of the specimens ranged from 13.73 to 37.3, with a median BMI of 23.92. $39\%$ of the specimens were female. All Specimens were acquired from Science Care Phoenix USA.

To obtain accurate pose information, we placed 14 stainless steel spherical fiducials within the frozen soft tissue of the specimens. These fiducials included seven with a 5 mm diameter and seven with a 3 mm diameter. Before inserting the fiducials, K-wires (3 mm diameter) were introduced to indicate our intended drilling trajectories. Fluoroscopic guidance was then used to verify the intended trajectories, and once confirmed, subsequent drilling was performed to create narrow tunnels within the frozen tissue for fiducial placement. After placing the fiducials, the residual tunnels were sealed with a low-viscosity superglue.

When inserting the fiducials, we ensured that the bony structures remained intact. Fiducials were strategically placed within an imaginary ellipsoid encompassing the spinal column, covering the area from L1 to L5. This careful positioning ensured that the fiducials had enough separation and were visible in clinically relevant imaging angles, including AP, lateral, and oblique views. 
Following the fiducial placement, we acquired a post-fiducial placement CT image of the specimen. This CT image explicitly depicted the relative placement of the fiducials with respect to the surrounding anatomy. The purpose of this post-placement CT image was to verify the accurate positioning of the fiducials and to measure their precise spatial coordinates in the 3D space of the CT volume.

After performing a 3D-3D registration \cite{johnson_brainsfit_2007} between the initial CT scan (moving image) and the post-fiducial placement CT scan (fixed image), a list containing the 3D fiducial coordinates $\mathbf{X}_i \in \mathbb{R}^3, i\in[1,2,...,14]$ was extracted from this CT image through initial thresholding of the metallic objects and later performing a 3D connected component analysis. The center of mass of the individual components corresponds to the object coordinates.  

While ensuring that the specimen remained frozen, we proceeded to collect intraoperative X-ray images using a clinical-grade mobile C-arm device (Cios Spin\textsuperscript{\textregistered}, Siemens Healthineers, Erlangen, Germany) with an effective detector size of 29.6 cm $\times$ 29.6 cm.

We initiated the image acquisition from the anterior-posterior (AP) view, which served as the base viewing angle. Subsequently, the C-arm gantry was incrementally rotated in the transverse plane, with $3^{\circ}$ increments, until a total orbit angle of $\pm102^{\circ}$ was achieved in both directions. Additionally, starting from the AP view angle, we utilized the C-arm's tilt movement in $3^{\circ}$ increments to image the specimen in the sagittal plane. The tilt angle was adjusted until a tilt of $\pm25^{\circ}$ was reached in both directions, resulting in the acquisition of oblique X-rays.

Similarly, we performed image acquisition from the lateral view angle, using the C-arm's tilt movement in $5^{\circ}$ increments in the sagittal plane. This allowed us to cover a tilt range of $\pm15^{\circ}$.
Throughout the image acquisition phase, we carefully positioned the C-arm gantry to ensure that the X-rays were centered at the L3 vertebra while capturing the entire lumbar region as completely as possible.

It is important to note that these precise angular increments were specific to the data collection phase to ensure a comprehensive dataset and are not required during the inference process. The small rotations down to 3° were manually executed to the best of our ability, utilizing the digital angles on the C-arm display as guides.

After acquiring the X-ray images, we performed image calibration to recover the extrinsic and intrinsic imaging parameters in a semi-automatic fashion using custom-developed software. The calibration process involved the use of 2D coordinates of the projected fiducials ($\mathbf{x}_j \in \mathbb{R}^2, j\in[1,2,...,14]$) and their corresponding 3D coordinates.

\begin{figure*}[t]
    \centering
    \includegraphics[width=\textwidth]{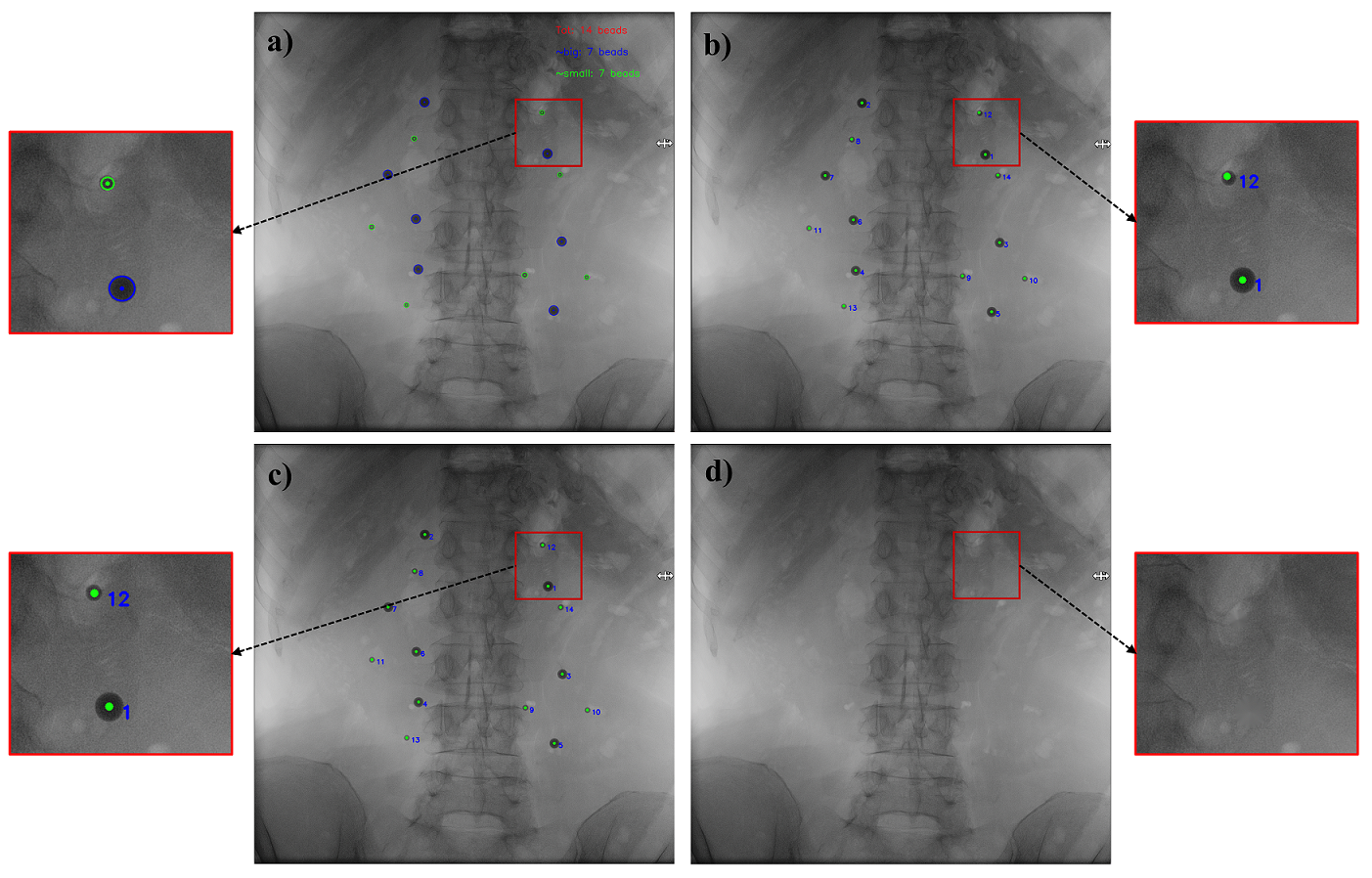}
    \caption{X-ray calibration pipeline. a) the fiducial projections are detected (blue: detected reference fiducial, green: detected fiducial), b) the initially estimated 2D-3D correspondence, c) the rectified 2D-3D correspondence, d) fiducial inpainting. }
    \label{fig:calib_pipeline}
\end{figure*}

The calibration process proceeded as follows:

\begin{itemize}
    \item \textbf{Fiducial Detection}: Initially, the 2D coordinates of the fiducials were detected using a circular Hough transformation algorithm \cite{duda_use_1972}, applied on the thresholded images. The software then displayed the detected fiducial points as suggestions to the user (Figure \ref{fig:calib_pipeline}-a) for verification or, if necessary, modification. 
    
    \item \textbf{Calculating Intrinsic and Extrinsic Parameters}: At this stage, the 2D and 3D fiducial coordinates were extracted. However, the correspondence between the two sets of coordinates ($\mathbf{x}_j \in \mathbb{R}^2$ and $\mathbf{X}_i \in \mathbb{R}^3$) was not established. To address this, we used seven fiducials with larger radii as reference fiducials, for which the corresponding 2D coordinates were desired in the first step. Assuming a random 2D-3D correspondence as the initial guess, we used the direct linear transform (DLT) algorithm \cite{abdel-aziz_direct_2015} to calculate the camera matrix $\mathsf{P}$ (containing the intrinsic and extrinsic imaging parameters) and the associated re-projection error. This re-projection error was used as the cost function in a random sample consensus (RANSAC) framework, whereby we repeated the calculation of the $\mathsf{P}$ matrix and the resultant re-projection error for all possible 2D-3D correspondences of the reference fiducials. The correspondence that yielded the smallest re-projection error was considered correct and was used to calculate $\tilde{\mathsf{P}}$, which was used as a preliminary estimate of the camera matrix (calculated using only the reference fiducials). The camera matrix $\tilde{\mathsf{P}}$ was then used to project all the fiducial points from 3D space $\mathbf{X}_i \in \mathbb{R}^3$ to 2D image space, represented as $\tilde{\mathbf{x}}_i = \mathsf{P}\mathbf{X}_i$, where $\tilde{\mathbf{x}}_i \in \mathbb{R}^2$ is the estimated 2D coordinates of the projected fiducials (Figure \ref{fig:calib_pipeline}-b).
    
    \item \textbf{Fiducial Coordinate Rectification}: By comparing the estimated fiducial projections $\tilde{\mathbf{x}}_i$ with the detected fiducial projections $\mathbf{x}_j$, we recovered the 2D-3D point correspondence for all fiducials. This allowed us to calculate the final estimate $\mathsf{P}$ by leveraging all fiducial markers (Figure \ref{fig:calib_pipeline}-c).
    
    \item \textbf{Fiducial Inpainting}: To preserve the realism of the calibrated X-rays, we inpainted the projections of the fiducial markers on the acquired X-rays using the methods reported in \cite{telea_image_2004} (Figure \ref{fig:calib_pipeline}-d).
\end{itemize}

After obtaining the camera matrix $\mathsf{P}$ from the calibration process, we decomposed it into the extrinsic and intrinsic camera parameters. The decomposition involved the following steps:

\begin{align}
    \mathsf{P} = \left[
\begin{array}{c;{2pt/2pt}c}
    \mathsf{KR} & -\mathsf{KR}\mathbf{X}_o 
\end{array}
\right] \rightarrow \mathsf{M} = \mathsf{KR},\quad \mathbf{m} = -\mathsf{KR}\mathbf{X}_o \\
    \mathbf{X}_o = -\mathsf{M}^{-1} \mathbf{m} \\
    \operatorname*{qr}(\mathsf{M}^{-1}) = \mathsf{R}^\intercal \mathsf{K}^{-1} \rightarrow q = \mathsf{R}^\intercal, \quad r = \mathsf{K}^{-1}
\end{align}

where $\mathsf{K}$ is the matrix containing the intrinsic camera parameters, $\mathsf{R}$ is the rotation matrix, $\mathbf{X}_o$ is the focal point coordinates, and $\operatorname*{qr}$ is used to decompose a matrix into an orthogonal matrix $q$ and an upper triangular matrix $r$. We generated the DRR image $I$ using Equation (\ref{eq:drr}) with the recovered intrinsic and extrinsic camera parameters.

Following this approach, the DRR image was projected from the same viewpoint as the acquired X-ray image, allowing us to obtain paired synthetic-real X-ray data.

We conducted the aforementioned process for all specimens, which included preoperative CT, fiducial placement, and post-fiducial CT. Subsequently, we calibrated and generated DRRs for all X-ray images in our \textit{ex-vivo} dataset. In this study, we had 13 \textit{ex-vivo} human specimens, resulting in 1017 paired synthetic-real X-ray images in our dataset. Utilizing the localization step, which will be introduced in Section \ref{sec:method:localization}, we cropped out individual lumbar vertebrae, which were fully visible.
This resulted in a total number of 4337 localized paired synthetic-real X-ray images. This dataset was split into 

\begin{itemize}
    \item \textbf{Train}: 10 \textit{ex-vivo} human specimen, 3317 images
    \item \textbf{Test}: 3 \textit{ex-vivo} human specimen, 1020 images.
\end{itemize}

Calibration errors can propagate into the DRR generation, leading to changes in perspective. We calculated the Euclidean distances for each pair of points to quantify the alignment between user-clicked 2D points on an X-ray and the corresponding projections of 3D fiducial coordinates. These distances served as a numerical measure of the paired data generation quality, capturing how well the acquired X-rays and the generated DRRs aligned. The results are presented in Section \ref{sec:results:paired}.

\subsection{Localization Network}
\label{sec:method:localization}
Before the reconstruction process, 2D localization and cropping of each lumbar vertebrae in the acquired X-ray image was required. To achieve this, we trained a YOLO v7 network \cite{wang_yolov7_2023} using both the synthetic and real datasets.

Using the CT segmentation labels from both real and synthetic datasets, we generated 2D masks through DRR generation. These masks were then employed to compute bounding boxes containing each vertebra. A 10\% margin around the maximum detected vertebra width was added to prevent unintentional clipping of important parts. The YOLO model, trained on these boxes, predicted the bounding boxes on new images. From the longer side of each predicted box, square patches containing the lumbar vertebrae of interest were cropped from both the original X-ray images and the corresponding DRRs.
The localization network was first trained on the DRRs. A version of it was then transfer-learned to work on real X-rays, allowing us to localize vertebrae within the target domain.

Due to the cropping process, a shift in the image plane is introduced, which can be described by the transformation matrix $\mathsf{Q}_i$ for each image $i$:
\begin{equation}
    \hat{\mathsf{P}}_i = \mathsf{Q}_i\mathsf{P}_i \: ,\quad \mathsf{Q}_i = \begin{bmatrix}
            1 & 0 & -t^x_i\\
            0 & 1 & -t^y_i \\
            0 & 0 & 1
        \end{bmatrix}.
    \label{eq:shift}
\end{equation}
Here, $\hat{\mathsf{P}}_i$ represents the adjusted P-matrix for the localized vertebra, and $t^x_i$ and $t^y_i$ are the horizontal and vertical shifts introduced during cropping.

In the productive setting, the surgeon can visually inspect localized vertebrae as a safety measure before passing them to the reconstruction network.

\subsection{3D Reconstruction Network}
\label{sec:method:network}
\begin{figure*}
    \centering
    \includegraphics[width=\textwidth]{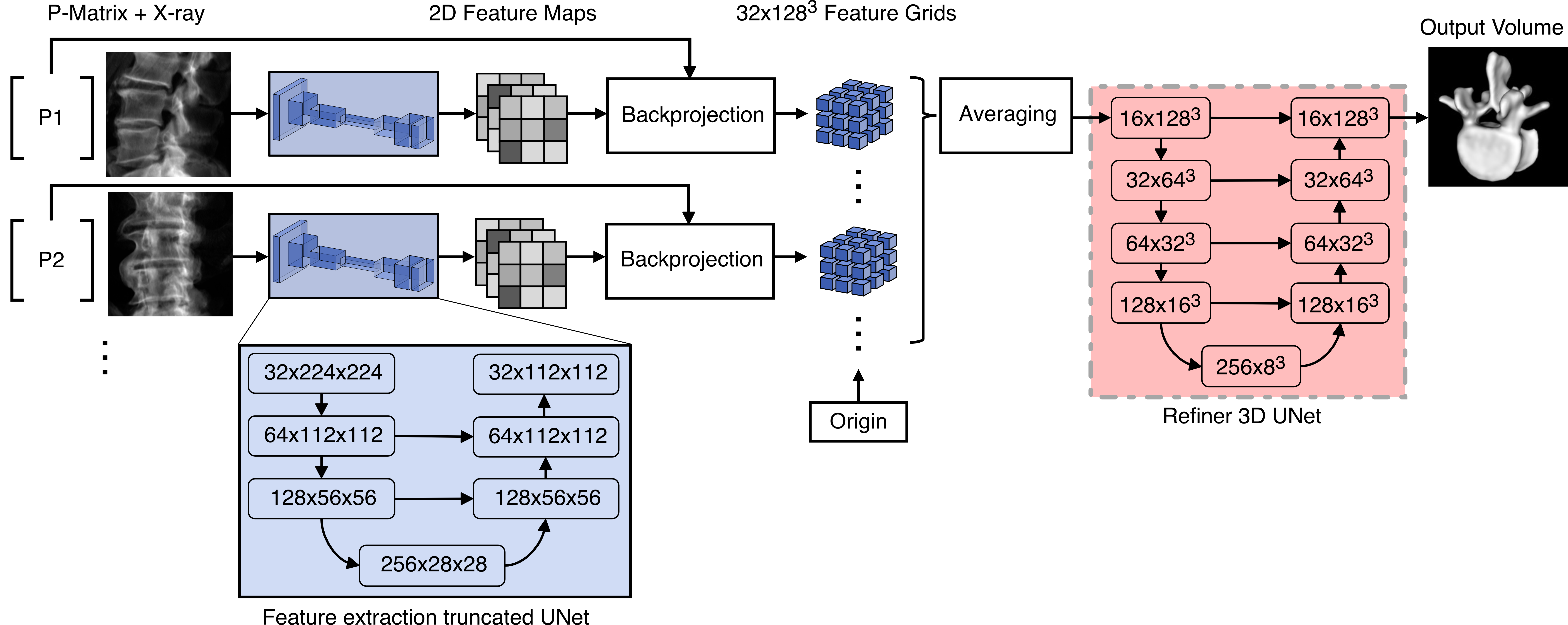}
    \caption{Our 3D reconstruction network. A 2D feature extraction stage first processes pairs of input images with the corresponding poses. These features are then back-projected into 3D, where they are averaged and refined by a 3D stage. The reconstruction is placed at a centroid estimated from the intersection of all input images.}
    \label{fig:network}
\end{figure*}

The centerpiece of our research is the 3D reconstruction network, which we introduced in our inaugural work \cite{jecklin_x23dintraoperative_2022}. The network, as depicted in Figure \ref{fig:network}, is designed to process a sparse set of input images and their corresponding poses containing extrinsic and intrinsic calibration information, represented by a projection matrix $\mathsf{P}$.

The input images of size $224\times224$ are initially processed by a 2D U-Net \cite{ronneberger_u-net_2015}, which extracts the essential features learned by the network for the 3D reconstruction of the vertebra. This process results in the generation of 32 feature maps of size $112\times112$. This enables the network to process vertebrae without the need for segmentation. The 32 2D feature maps are then back-projected into 3D, utilizing the adjusted projection matrices $\hat{\mathsf{P}}_i$. This results in a $128\times128\times128$ 3D feature grid for each input image. These feature grids allow the network to process an arbitrary number of input image-pose pairs.
Following the feature grid aggregation, a 3D U-Net is tasked with refining the grid and generating a 3D reconstruction. 
Due to the utilization of projection matrices, the features intersect in the correct spatial location. However, the 3D feature grid must be appropriately positioned to capture those intersections and fully contain the resulting vertebra. Therefore, providing an origin for the $128^3$ reconstruction cubes is essential.
The synthetic data's origin is known. When dealing with real X-rays, an estimation of the reconstruction origin is necessary. 

It is important to note that the entire process is end-to-end trained, which results in intermediate representations, both before and after backprojection, that are not easily humanly interpretable. 
An alternative implementation could use a separate segmentation network to isolate vertebrae before backprojection, followed by a distinct refiner network, to achieve human-interpretable intermediate representations. However, this approach prevents the utilization of information from multiple views to improve both vertebrae isolation and reconstruction. Errors in the segmentation could directly propagate into the reconstruction, leading to inaccuracies. Such errors can arise due to the varying difficulty of segmenting vertebrae, influenced by factors such as the pose of X-ray acquisition, human anatomical variations, and occlusion,e.g., through organs. Our approach leverages information from multiple views throughout the entire process, providing a comprehensive method for 3D reconstruction.

\begin{figure}
    \centering
    \includegraphics[width=\linewidth]{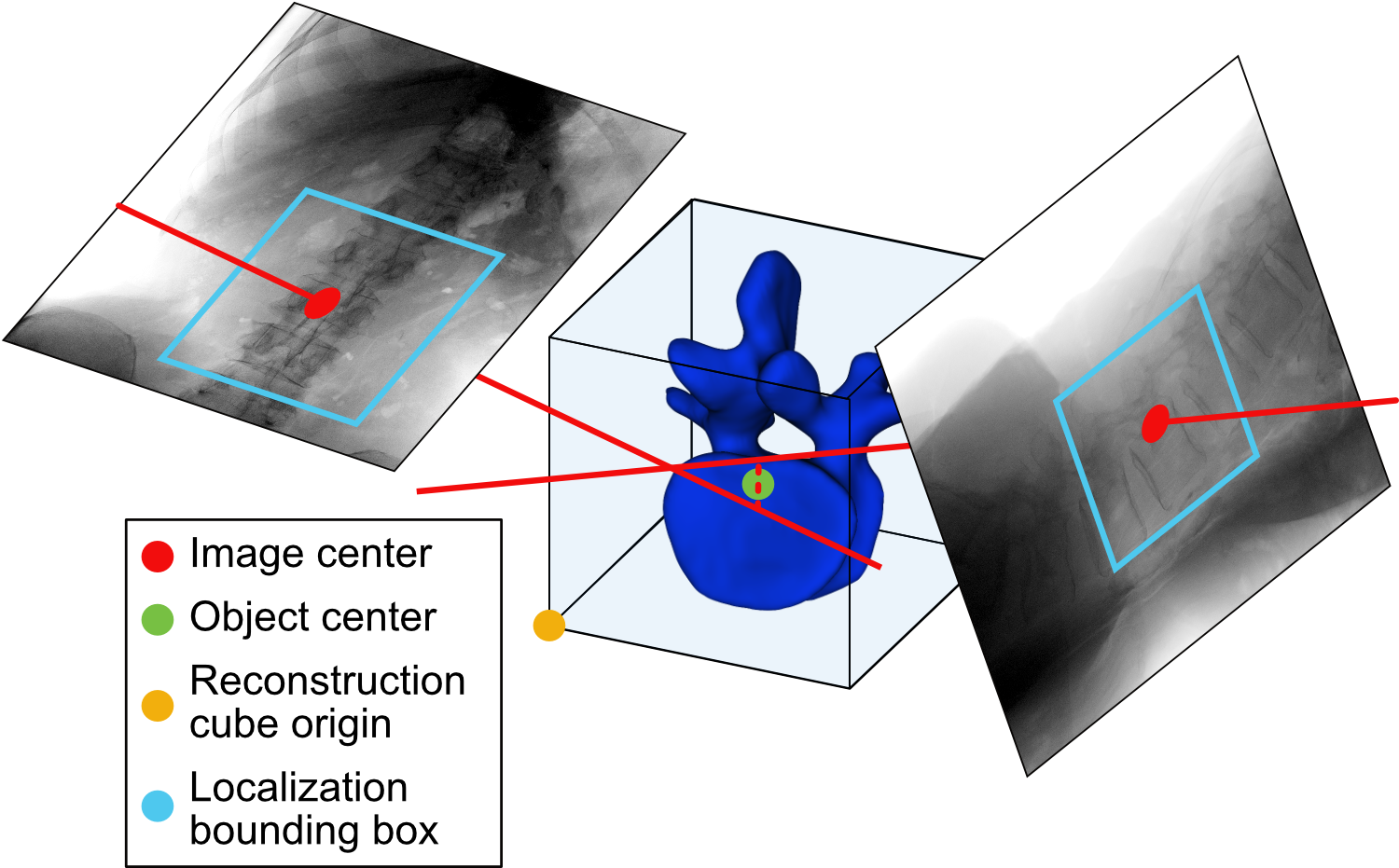}
    \caption{Visual representation of the reconstruction cube origin estimation. A vertebra level of interest is first localized in all images, resulting in a shift addressed with Equation (\ref{eq:shift}). The closest intersection of the back-projected vertebra center through the 2D image centers is calculated (Equation (\ref{eq:intersection})). The resulting object center has a known offset from the reconstruction cube origin, which is crucial for accurate 3D reconstruction.}
    \label{fig:origin}
\end{figure}
Given the localization of individual vertebrae, we can assume that the resulting cropped images were centered approximately at the center of the vertebra to be reconstructed. Therefore, the optical axes of the pinhole cameras corresponding to $\hat{\mathsf{P}}_i$ intersect within the vertebra. The intersection can be computed using a variety of methods. We used the commonly used linear triangulation \cite{hartley_multiple_2003}, which finds an approximate solution in the least-squares sense. We solve a linear system of the form $\mathsf{A}\mathbf{X} = 0$ , where $\mathsf{A}$ is
\begin{equation}
    \mathsf{A}=\left[\begin{array}{c}
    u_1 \hat{p}_1^3-\hat{p}_1^1 \\
    v_1 \hat{p}_1^3-\hat{p}_1^2 \\
    \vdots \\
    u_n \hat{p}_n^3-\hat{p}_n^1 \\
    v_n \hat{p}_n^3-\hat{p}_n^2
    \end{array}\right]
    \label{eq:intersection}
\end{equation}

and $[u_i,v_i]$ denotes the center of the localized images of the $i$th camera. $\hat{p}^j_i$ denotes the $j$th row and $i$th column of the projection matrix $\hat{\mathsf{P}}_i$. We received $\mathbf{X}$ by solving this linear equation system using singular value decomposition \cite{hartley_multiple_2003}.

The estimation of this origin is visualized as an example using two images in Figure \ref{fig:origin}.

Compared to the network presented in \cite{jecklin_x23dintraoperative_2022}, we made two changes. Firstly, we replaced the 2D interpolation, previously employed for back-projecting 2D features into 3D, with a nearest-neighbor approach.
After extensive testing, we identified that this change results in faster training times without compromising reconstruction quality.
Secondly, we opted for a new loss function, combining the dice loss and cross-entropy loss (DiceCE loss) instead of using the cross-entropy loss alone. In doing so, we achieved faster convergence during the training phase. Contrary to common intuition, the 3D reconstructed vertebrae only occupy about 3\% to 6\% of the $128^3$ reconstruction volume, leading to heavily imbalanced labels. The introduction of the DiceCE loss effectively addressed this issue and significantly contributed to performance improvement.

\subsection{In-Silico Training}
\label{sec:methods:insilico}
As discussed earlier, our network can be trained on synthetic data (DRRs) or real X-rays. Given the aforementioned limited availability of labeled real data, the model as described in Section \ref{sec:method:network} underwent initial training on synthetic data introduced in Section \ref{sec:method:synthetic_data}. Hence, we refer to the network trained exclusively on synthetic data as the \textit{base model}, while the model subsequently trained on real X-rays is termed the \textit{transfer-learned model}. Details of the training parameters are presented in Section \ref{sec:methods:implementation}.

\subsection{Domain Adaptation}
\label{sec:methods:domad}
\subsubsection{Transfer Learning}
\label{sec:methods:domad:tl}
Following this initial training, the network underwent transfer learning on the real X-ray dataset composed as described in Section \ref{sec:method:real_data}.

We conducted tests during the transfer learning experiments by freezing different network modules, such as the initial or output layers. However, as our experiments demonstrated that the network achieved the best performance when all layers were trainable during the transfer learning phase, we consistently maintained this approach.

\subsubsection{Style Transfer}
\label{sec:methods:st}
In addition to the abovementioned \textit{transfer-learned model}, we employed a style transfer paradigm for bridging the gap between the X-ray source domain and the synthetic DRR domain, which was the training ground for our \textit{base model}. Opting for the reverse direction, in contrast to adapting synthetic data to the X-ray domain, serves two previously outlined purposes. 
The first benefit of our chosen style transfer direction is that it transforms X-ray images into an inevitably simplified synthetic representation (DRR). Conversely, reversing this process would require the network to generate anatomical features based on its learned training distribution. 
Secondly, this direction allows applying both existing and future synthetically trained networks to real X-ray images, enhancing model consistency and generalizability. 
Transfer-learning, using synthetic data adapted to the X-ray domain, would require retraining the reconstruction model each time the target real domain needs expansion, such as when incorporating data from new types of C-arms or different imaging conditions. This approach significantly increases the complexity and resource requirements for maintaining and updating the model. Moreover, relying solely on the collected X-rays would not provide enough data to train such a large network to convergence, highlighting the practicality of our method. Style transfer to the controlled synthetic environment offers a more standardized platform for training, leading to improved feature learning and resource efficiency. To achieve this, we utilized a cGAN network architecture known as Pix2Pix \cite{isola_image--image_2018}. The key components of this architecture include

\begin{itemize}
    \item \textbf{Generator}: The generator network takes an image from the source domain as input and aims to produce a realistic image that resembles the target domain. It achieves this by employing an adversarial loss, a characteristic feature of Generative Adversarial Networks (GANs) \cite{goodfellow_generative_2014}. This loss encourages the generator to create images that are indistinguishable from real target domain images.
    \item \textbf{Discriminator}: The discriminator network is crucial in our style transfer approach. It receives the generator's output and a real image from the target domain as inputs. The discriminator's primary task is distinguishing between the generated and target domain images. The discriminator guides the training process by providing feedback to the generator to produce more realistic and accurate translations.
    \item \textbf{L1 Loss}: In addition to the adversarial loss, the style transfer model incorporates an L1 loss. This loss function measures the pixel-wise difference between the generated image and the expected output image (i.e., the real target image). The L1 loss encourages the generator to produce realistic images and exhibit visual similarity to the target domain. This ensures that fine details and structural characteristics are faithfully preserved.
\end{itemize}

When trained with the paired dataset, the style transfer model excels at image-to-image translation tasks using both adversarial and L1 losses.

\subsection{Experiments and Performance Evaluation}
\label{sec:methods:performance_eval}
\begin{figure*}[t]
    \centering
    \includegraphics[width=\linewidth]{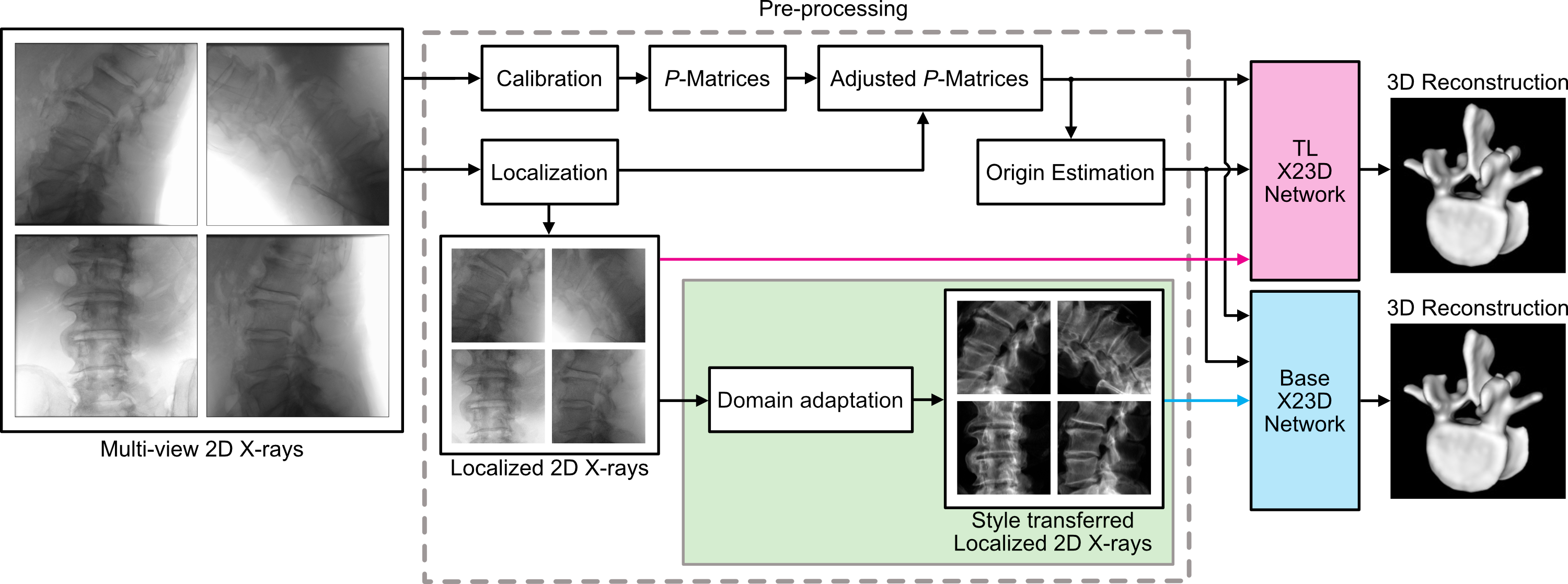}
    \caption{Pipeline for the inference of our 3D reconstruction approach. Input X-rays undergo a calibration stage to generate P-matrices. These adjusted P-matrices incorporate translations from the localization step. Using these matrices, we estimate the origin of the 3D reconstruction cube. The X23D network utilizes the estimated origin, adjusted P-matrices, and the corresponding images to create a 3D model. X-rays can either be directly processed by a \textit{transfer-learned model} (magenta path) or transformed to the DRR domain (green) before processing with our \textit{base model} (cyan path).}
    \label{fig:inference}
\end{figure*}
We evaluated the compound 3D reconstruction accuracy of the entire X23D pipeline based on the unseen images of the paired test dataset. The evaluation was performed on both synthetic and real datasets, depending on the type of ablation study.

To thoroughly evaluate all stages of our pipeline, including localization, domain adaptation, and 3D reconstruction, we employed a train-test split at the level of individual \textit{ex-vivo} human specimens as outlined in \ref{sec:method:real_data}. This approach was adopted to prevent any potential information leakage or bias that might otherwise distort the results.

To further validate the reconstruction accuracy and enhance the robustness of our findings, we conducted additional evaluations on six previously unseen specimens. The imaging data for these specimens were acquired using the same Siemens Cios Spin device as in the initial data acquisition. Additionally, we utilized a Ziehm Vision system with a 21x21 cm detector and a Ziehm Vision FD system (Nuernberg, Germany) with a 31x31 cm detector to diversify the imaging conditions and ensure comprehensive assessment. For each specimen, we collected multiple sets of four X-rays, with each set consisting of images acquired from roughly AP, lateral, oblique, and a miscellaneous pose. These images were then used to reconstruct the visible vertebrae covered in the X-rays.

Figure \ref{fig:inference} shows the paths data takes through our pipeline during inference. The images passed to our network undergo pre-processing, including calibration. 
Unlike the calibration setup detailed in the data generation process (Section \ref{sec:method:real_data}), absolute poses relative to the CT are not required for inference; only the relative poses between images are needed. For our experiments on \textit{ex-vivo} human specimens, we placed a radiolucent object containing 14 radiopaque metal beads with a known geometry next to the human specimen. This phantom aids in establishing the relative poses of the input X-ray images through the perspective n-points (PnP) algorithm. The same approach as detailed in \ref{sec:method:real_data} was utilized to inpaint the metal beads from the X-rays.
After the pre-processing stage, the data can follow one of two paths. The first option is to pass the localized X-rays directly to the \textit{transfer-learned model}. In this approach, the network can directly generate the 3D reconstruction from the localized X-ray images.
The second option involves passing the localized X-rays to the trained style-transfer network. The style-transfer network transforms the localized X-ray images into the DRR domain, where they can be further processed by the \textit{base model}, which was entirely trained using synthetic data.

We conducted a series of ablation studies to assess the robustness and sensitivity of our network to varying numbers and poses of input images. This involved comparing our predictions to the ground truth segmented CT scans, using synthetic data with the \textit{base model}.

In a separate ablation study, we targeted challenging areas on the vertebra for reconstruction. For this, we utilized the real dataset and the \textit{base model} in combination with the style transfer for domain adaptation.

Lastly, we conducted an ablation study to understand the impact of using separately trained models for each vertebral level versus our all-in-one \textit{base model} using synthetic data and the \textit{base model}. 

We used the following metrics to assess the performance of the 3D reconstructions: F1 score, Intersection over Union (IoU), surface score \cite{knapitsch_tanks_2017,tatarchenko_what_2019}, Hausdorff distance with a 95\% percentile (HD95) and average surface distance (ASD). ASD and HD95 are reported in $mm$. 

The F1 score is a commonly used metric that evaluates the overall volume agreement between predicted and ground-truth 3D reconstructions. Although it provides a measure of overlap, it may not fully capture the fine surface details crucial in the context of intraoperative surgical guidance.

Therefore, we used the surface score metric as a specific evaluation of the predicted surfaces compared to the ground truth. The surface score measures the accuracy of the reconstructed surface geometry by calculating the average distance between the predicted surface points and the corresponding points on the ground truth surface. This metric is particularly useful for assessing the fine details and intricacies of the vertebral surfaces, which are essential for precise intraoperative guidance. Unlike the F1 score, which focuses on volume overlap, the surface score provides a more detailed evaluation of surface accuracy, ensuring that our 3D reconstructions accurately represent the individual vertebrae.

Section \ref{sec:method:surface_heatmap} illustrates the impact of patient-specific reconstruction on both the surface and F1 scores. 

\subsubsection{Reconstruction Performance on Real X-rays}
In this section, we comprehensively assessed the performance of the entire pipeline on real X-rays. 

We established a baseline using a naive approach, where the \textit{in-silico} trained base network was directly fed with the collected real X-ray images. Subsequently, we evaluated both the \textit{transfer-learned model}, and the \textit{base model} using domain-adapted X-ray data. 

To comprehensively assess our method's performance, we present scores averaged across 100 reconstructions of 15 vertebrae from three specimens of the \textit{ex-vivo} test set. For the reconstructions, four images from the view angles priorly defined as AP, lateral oblique, and miscellaneous have been chosen according to the findings in section \ref{sec:methods:num_of_views} and \ref{sec:methods:view_combination}.

We explore the distribution of prediction quality by plotting a histogram of resulting surface scores. This analysis provides detailed insight into our methods' overall performance characteristics.

Two ablation studies were performed. To explore the full potential of our network, we conducted an experiment utilizing the available paired dataset, providing the base network with DRRs corresponding to the poses of real X-rays. This experiment provided an upper limit for achievable scores as it bypassed the domain gap by directly feeding the base network with synthetic images on which it was trained. 

To assess the influence of origin estimation accuracy, we compared it with the outcomes achieved using the synthetic dataset's known ground truth origin.

We conducted the experiment multiple times to determine the inference time of the 3D reconstruction and localization models. The average time for the 3D reconstruction was derived from 429 vertebrae reconstructions using four input images each. Additionally, the inference time for the localization model was ascertained using a dataset of 220 real X-ray images.

\subsubsection{Number of Images}
\label{sec:methods:num_of_views}
To determine the ideal number of sparse images for accurate 3D reconstruction, we performed ablation experiments on synthetic data, starting with standard clinical views, namely AP and lateral. We then progressively added oblique and miscellaneous views, as defined in Section \ref{sec:method:synthetic_data}. For each experiment, we evaluated the performance averaged across the entire test set, performing a reconstruction with the specified number and combination of views. Table \ref{tab:number_of_views} shows an overview of these combinations.
\begin{table}[ht]
\centering
\begin{tabular}{ccccc}
\hline
\textbf{\#} & \textbf{AP} & \textbf{Lateral} & \textbf{Oblique} & \textbf{Misc} \\
\hline
2 & 1 & 1 & 0 & 0 \\
3 & 1 & 1 & 1 & 0 \\
4 & 1 & 1 & 1 & 1 \\
5 & 2 & 1 & 1 & 1 \\
6 & 2 & 2 & 1 & 1 \\
7 & 2 & 2 & 2 & 1 \\
8 & 2 & 2 & 2 & 2 \\
\hline
\end{tabular}
\caption{Different view combinations used to evaluate the optimal number of sparse images for the 3D reconstruction task.}
\label{tab:number_of_views}
\end{table}

\subsubsection{View Angle Combination}
\label{sec:methods:view_combination}
In this test, we evaluated the impact of different combinations of view angles for a given number of input images using our synthetic dataset. Although there is interdependence between view combinations and the ideal number of images, this evaluation primarily focused on the importance of specific view categories. Preliminary results suggested that adding more views, even when spatially close to existing views, can enhance the quality of the reconstruction. Thus, we explored whether having multiple AP and lateral views may provide more valuable information than including various view angles, such as oblique and miscellaneous views.

\subsubsection{Sensitivy to View Angles}
We investigated the impact of deviating from the standard 28 poses used in the synthetic dataset to assess the network's robustness to varying view angles. In this test, we fixed the best number and combinations of input images from previous experiments, except for one view angle. For this view, we varied the angle in two directions on a grid pattern. The grid was made of $21 \times 21$ points with a maximum of $20^{\circ}$ deviation in both directions. The resulting surface scores on that closely sampled grid were then interpolated, resulting in a 2D heatmap. This heatmap enabled us to visualize how the network's performance changed with slightly different viewpoints, providing insight into its ability to handle variations encountered during intraoperative X-ray acquisition. The experiment used a vertebra randomly selected from a CT scan within the test set.

\subsubsection{Challenging Regions}
\label{sec:method:surface_heatmap}
Accurate 3D reconstruction faces greater challenges in specific anatomical regions, such as transverse and spinous processes, than in the vertebral body. The extent of these challenges primarily depends on the selected views. To identify and visualize these challenging areas systematically, we generated 3D distance maps that represent varying levels of reconstruction performance.

We conducted 100 reconstructions using X-rays from the \textit{ex-vivo} test set, following the optimal view angle combinations identified in the experiment (Section \ref{sec:results:combination}). The resulting reconstructions were sorted according to their surface scores, and we selected predictions corresponding to surface scores of $60\%$, $70\%$, $80\%$, and $90\%$. These thresholds were chosen to encompass a range of reconstruction performances.

In defining accuracy for our approach, we consider the clinical requirements, particularly for procedures such as pedicle screw placement. In our experience, achieving reconstruction accuracy with deviations smaller than 1 mm is generally necessary for clinical utility.

\subsubsection{Level-specific Models}
We conducted experiments with models trained solely on one vertebra to test whether individually trained models can outperform a combined model. For this, we averaged scores on a vertebral level over 100 random but clinically feasible combinations of four synthetic images. 

\subsection{Coding Platform, Hardware, and Training Hyperparameters}
\label{sec:methods:implementation}
In the development of the DRR creation software, we utilized the Insight Toolkit (ITK) framework \cite{mccormick_itk_2014} and the SimpleITK framework \cite{lowekamp_design_2013}. Additionally, we employed PySide6 and QT to create the software's graphical user interface (GUI).

We use the PyTorch framework to build and train the 3D reconstruction network \cite{paszke_pytorch_2019}. Training processes and tests were performed on Nvidia RTX 3060 with 8 GB of VRAM each.

Training of the base network on synthetic data required around 40 epochs. We used an Adam optimizer with a learning rate of $10^{-4}$ and a weight decay of 0.9 during the training process. The transfer learning on real X-rays also required approximately 40 Epochs. The learning rate was reset to the initial value of $10^{-4}$. Early stopping was applied as a regularization technique to avoid overfitting.

We utilized the Python implementation of Pix2Pix \cite{zhu_cyclegan_2023} to perform the style transfer from X-ray to DRRs. The implementation offers flexibility in tuning the hyperparameters, including the input shape and the number of channels. Although our dataset is grayscale, we configured the model using three channels, as discussed in Sections \ref{sec:method:synthetic_data} and \ref{sec:method:real_data}. The network was trained for 400 epochs.

\section{Results}
\label{sec:results}
\subsection{Reconstruction Performance on Real X-rays}
\label{sec:results:reconstruction}
\begin{table*}[htbp]
    \centering
    \begin{tabular}{p{2cm}cccccc}
        \textbf{Origin} & \textbf{Method} & \textbf{Surface$\uparrow$} & \textbf{F1$\uparrow$} & \textbf{IoU$\uparrow$} & \textbf{ASD$\downarrow$} & \textbf{HD95$\downarrow$} \\
        \hline
        \multirow{3}{*}{estimated} 
        & ST & 0.62 $\pm$ 0.08 & 0.84 $\pm$ 0.03 & 0.86 $\pm$ 0.02 & 0.97 $\pm$ 0.20 & 3.68 $\pm$ 1.10 \\
        & TL & 0.63 $\pm$ 0.10 & 0.84 $\pm$ 0.04 & 0.86 $\pm$ 0.02 & 0.98 $\pm$ 0.23 & 3.45 $\pm$ 0.90 \\
        & DRR & 0.69 $\pm$ 0.07 & 0.86 $\pm$ 0.03 & 0.88 $\pm$ 0.02 & 0.84 $\pm$ 0.16 & 3.33 $\pm$ 1.38 \\
        \hline
        \multirow{3}{*}{ground truth} 
        & ST & 0.73 $\pm$ 0.10 & 0.88 $\pm$ 0.04 & 0.89 $\pm$ 0.03 & 0.73 $\pm$ 0.20 & 3.01 $\pm$ 2.21 \\
        & TL & 0.66 $\pm$ 0.12 & 0.85 $\pm$ 0.08 & 0.87 $\pm$ 0.04 & 0.89 $\pm$ 0.25 & 3.17 $\pm$ 1.67 \\
        & DRR & 0.79 $\pm$ 0.11 & 0.89 $\pm$ 0.06 & 0.90 $\pm$ 0.04 & 0.63 $\pm$ 0.24 & 2.86 $\pm$ 3.72 \\
        \hline
    \end{tabular}
    \caption{Results on Real Data: The experiment was conducted twice, once with estimated reconstruction cube origins and once with origins from the ground truth volumes. In the Style-Transfer method (ST), the real X-rays from the paired test set were first processed by the style transfer network before being passed to the \textit{base model}. In the transfer-learned (TL) method, the \textit{transfer-learned model} processed the X-rays from the paired test set. In the DRR method, the \textit{base model} was provided with DRRs from the paired test set.}
    \label{tab:real_data}
\end{table*}

Table \ref{tab:real_data} presents the 3D reconstruction performance using data from the paired test set. The first two rows show the results of the style transfer (ST) and transfer-learned (TL) approach. Both process real X-rays. The former approach uses style transfer for domain adaptation before passing the data to the \textit{base model}. The latter approach uses the \textit{transfer learned model}. The third row (DRR) shows the result of the \textit{base model} provided with the synthetic images of the paired dataset.

Each experiment was conducted twice, using both the estimated reconstruction cube origin and the origin from the ground truth volume. The results are all averaged over a hundred reconstructions based on different view combinations for each vertebra. 

With estimated origins, $62\%$ and $63\%$ surface scores are achieved for the style transfer and transfer-learning approach, respectively. 

When using the ground truth origin, the \textit{transfer-learned model} improved by $3\%$, without a notable difference in the F1 score. The style transfer approach utilizing the ground truth origins achieved a $73\%$ surface and a $88\%$ F1 score. This corresponds to a $11\%$ and $4\%$ boost, respectively. This makes it perform equally well on real data as our previous work \cite{jecklin_x23dintraoperative_2022} on entirely synthetic data. All observed surface scores and F1 values using estimated origins significantly differed from their ground truth counterparts. The highest p-value, $1.5892 \times 10^{-5}$, was observed between the F1 scores of the \textit{transfer-learned model} groups.

Compared to our previous work, we use more synthetic data to train our \textit{base model}. DRRs and ground truth origins provided to the retrained \textit{base model} achieves a $79\%$ surface score while maintaining the same F1 score as previously reported in \cite{jecklin_x23dintraoperative_2022}. The surface score is $10\%$ higher than the DRR approach using estimated origins.

\begin{figure}
    \centering
    \includegraphics[width=\linewidth]{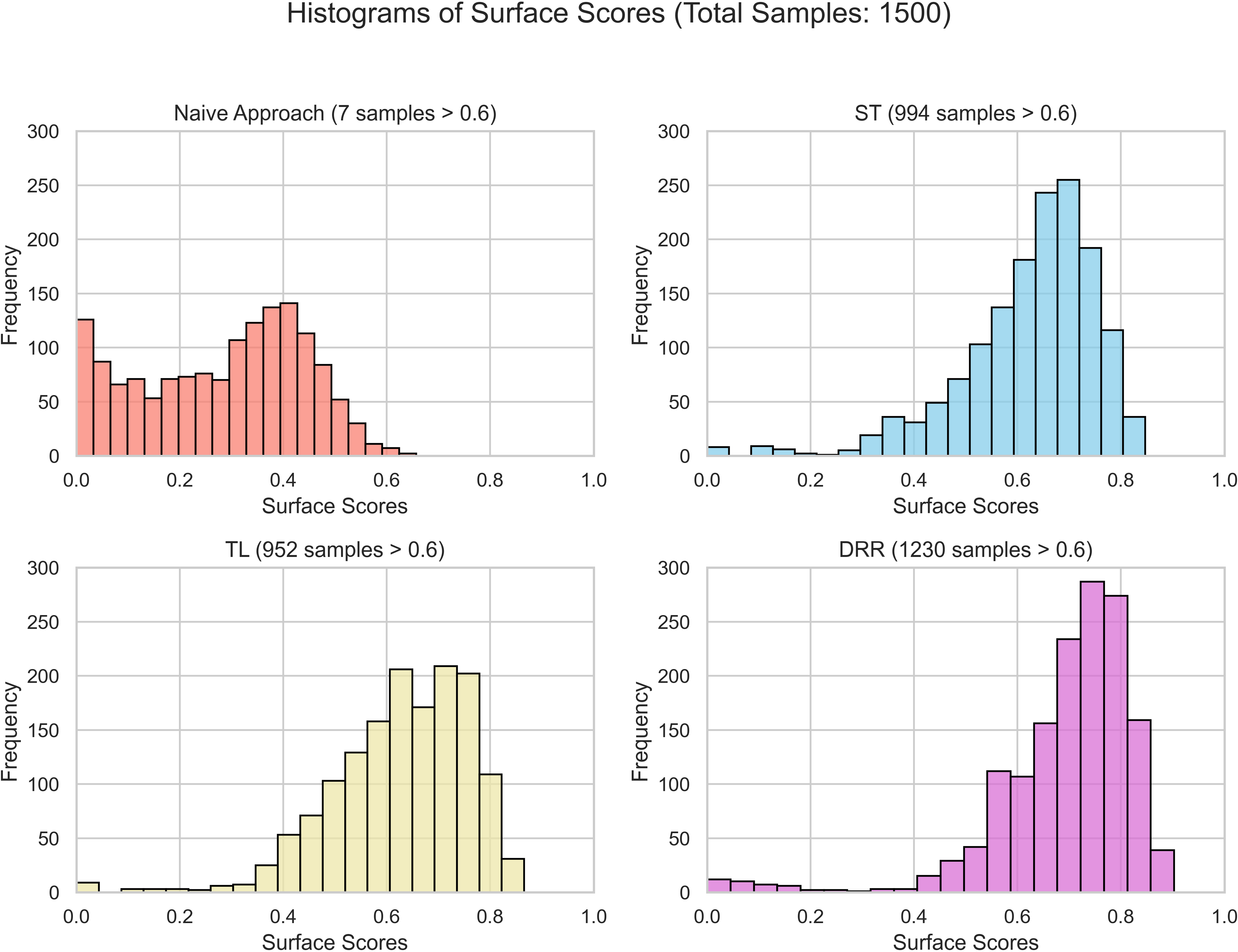}
    \caption{Histograms depicting surface scores derived from reconstructions using the four methods (Naive Approach, ST, TL, and DRR) with an estimated centroid. Each histogram comprises 1500 scores, resulting from 100 reconstructions of five vertebrae from three \textit{ex-vivo} human specimens in the paired test set.}
    \label{fig:hist}
\end{figure}
Figure \ref{fig:hist} displays histograms representing the surface scores for the 100 reconstructions of each vertebra in the paired test set. The origin was estimated for those reconstructions. There are four histograms, one for each approach outlined above. Additionally, we display one histogram for the naive approach where the \textit{base model} was provided with X-rays. 
The performance of the naive approach was expectingly low. Almost no sample achieved a surface score greater than 0.6. On the other hand, the \textit{base model} provided with DRRs performs well in most cases, with most surface scores exceeding 0.6.

\begin{figure*}
    \centering
    \includegraphics[width=\linewidth]{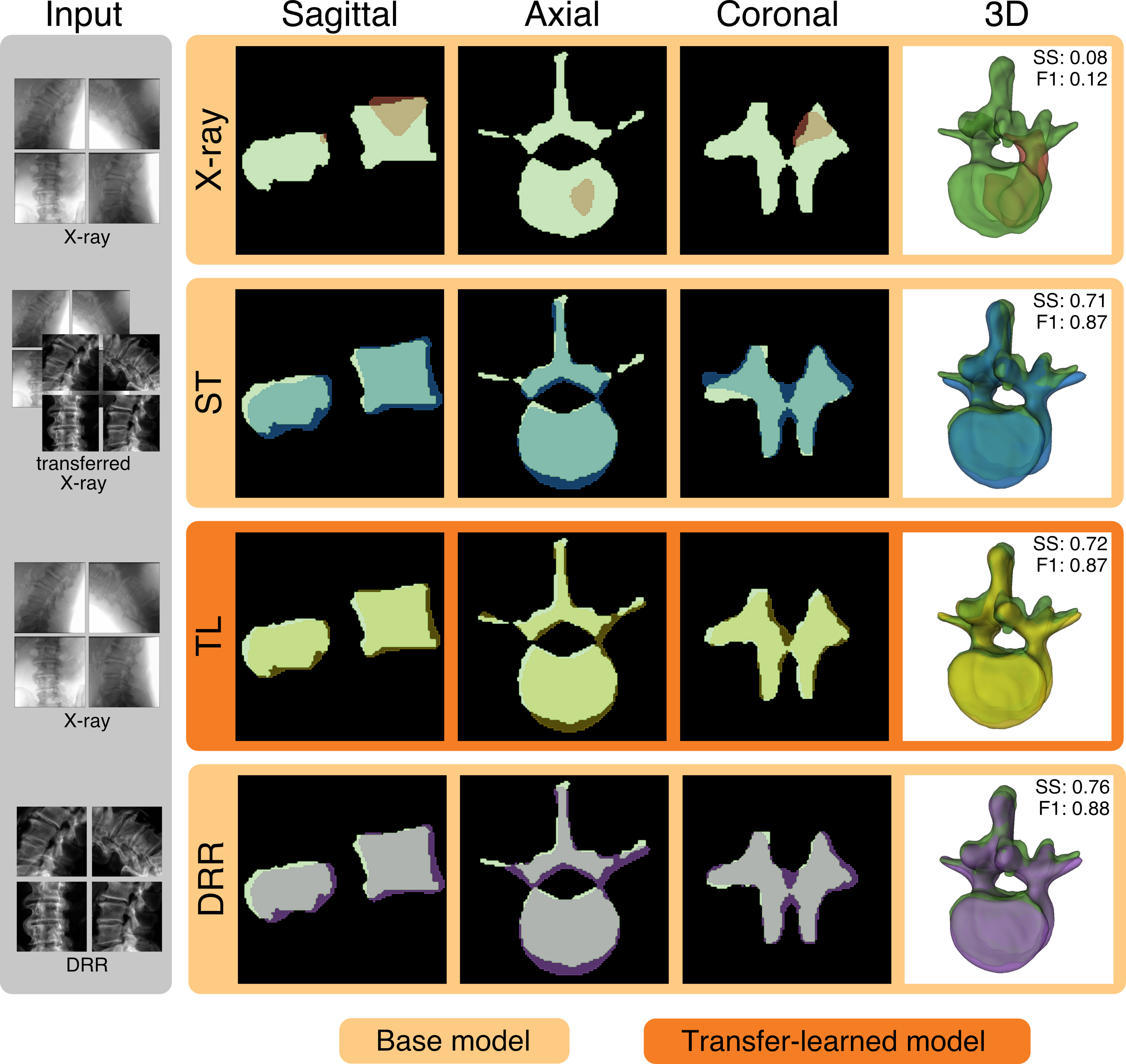}
    \caption{The figure showcases 3D reconstructions using our complete pipeline on unseen data, with estimated origins. Each column represents a different perspective, including sagittal, axial, and coronal planes, and a 3D view, enabling a comprehensive assessment of the reconstructions. The ground truth is depicted in green. The first row presents the naive approach using X-rays with the \textit{base model}. The second row shows real X-rays transformed into the DRR domain and processed by the \textit{base model}. The third row displays results from our \textit{transfer-learned model} on real X-rays. The fourth row demonstrates DRRs generated from the same X-ray viewpoints processed by the \textit{base model}, which acts as a theoretical upper limit. The last column also shows the surface score (SS) and F1 score.}
    \label{fig:4_comparison}
\end{figure*}

Figure \ref{fig:4_comparison} shows one particular reconstruction using these four methods. Each method received the same randomly picked four images from the paired test set in the required image modality (X-ray or DRR). The naive approach resulted in the lowest overall surface score of $8\%$. The theoretical upper limit of the \textit{base model} is achieved when providing it with DRRs. This resulted in $76\%$ surface score and $88\%$ F1 score. Comparable F1 and surface score results were achieved using the style-transfer network or the transfer-learned approach. Namely $71\%$ and $87\%$ F1 and $72\%$ and $87\%$ surface score, respectively. Surface scores were $6.6\%$ and $5.3\%$ lower compared to the model provided with DRRs.

\begin{table*}[htbp]
    \centering
    \begin{tabular}{c|ccccccc}
        \textbf{Specimen} & \textbf{C-Arm} & \textbf{n} & \textbf{Surface$\uparrow$} & \textbf{F1$\uparrow$} & \textbf{IoU$\uparrow$} & \textbf{ASD$\downarrow$} & \textbf{HD95$\downarrow$} \\
        \hline
        1 & Siemens & 10 & 0.68 $\pm$ 0.08 & 0.86 $\pm$ 0.03 & 0.75 $\pm$ 0.05 & 0.89 $\pm$ 0.17 & 3.55 $\pm$ 1.50 \\ \hline
        2 & Siemens & 8  & 0.64 $\pm$ 0.07 & 0.85 $\pm$ 0.02 & 0.75 $\pm$ 0.03 & 0.93 $\pm$ 0.15 & 4.39 $\pm$ 0.97 \\  \hline
        \multirow{2}{*}{3} & Ziehm31x31 & 5 & 0.65 $\pm$ 0.10 & 0.84 $\pm$ 0.05 & 0.72 $\pm$ 0.07 & 0.89 $\pm$ 0.24 & 5.00 $\pm$ 2.08 \\
        & Siemens & 5 & 0.71 $\pm$ 0.04 & 0.86 $\pm$ 0.01 & 0.76 $\pm$ 0.02 & 0.73 $\pm$ 0.05 & 4.14 $\pm$ 0.81 \\ \hline
        \multirow{2}{*}{4} & Ziehm21x21 & 2 & 0.73 $\pm$ 0.02 & 0.88 $\pm$ 0.01 & 0.79 $\pm$ 0.01 & 0.80 $\pm$ 0.07 & 2.45 $\pm$ 0.25 \\
        & Siemens & 4 & 0.65 $\pm$ 0.03 & 0.86 $\pm$ 0.01 & 0.75 $\pm$ 0.02 & 1.03 $\pm$ 0.16 & 3.02 $\pm$ 0.47 \\ \hline
        5 & Siemens & 5 & 0.69 $\pm$ 0.07 & 0.85 $\pm$ 0.03 & 0.74 $\pm$ 0.04 & 0.91 $\pm$ 0.18 & 4.38 $\pm$ 2.83 \\ \hline
        \multirow{2}{*}{6} & Ziehm21x21 & 3 & 0.73 $\pm$ 0.04 & 0.85 $\pm$ 0.02 & 0.74 $\pm$ 0.02 & 0.72 $\pm$ 0.08 & 3.40 $\pm$ 0.78 \\
        & Siemens & 5 & 0.72 $\pm$ 0.06 & 0.85 $\pm$ 0.02 & 0.74 $\pm$ 0.03 & 0.77 $\pm$ 0.08 & 3.15 $\pm$ 0.78 \\
        \hline
    \end{tabular}
    \caption{Summary of reconstruction performance for various experiments, each identified by a unique specimen number. The table presents the number of vertebrae reconstructed (n) alongside performance metrics for different C-Arm systems.}
    \label{tab:various_results}
\end{table*}

Table \ref{tab:various_results} shows the result of our method on six unseens specimens using the style-transfer approach. Both surface score and F1 score are above the reported range using estimated origins 
of Table \ref{tab:real_data}. The results using images taken with the Ziehm C-arms with smaller and larger sensors are in the same range as the results achieved with Siemens C-arms. A t-test was performed to compare the reconstruction accuracy between the Siemens and Ziehm C-arms. The results indicated no significant difference in accuracy between the different C-arms. However, it is important to note that this does not conclusively prove that there is no difference. The sample size is not sufficient for a test such as the Kolmogorov-Smirnov test. Thus, while preliminary results suggest comparable performance across the different C-arms, further studies with larger sample sizes are necessary to confirm these findings.

The average reconstruction time without localization for 429 test samples was $81.1$ ms. During the inference process, the localization step took an average of $10.3$ ms across 220 images.

\subsection{Number of Images}
\label{sec:results:num_of_images}
\begin{figure}
    \centering
    \includegraphics[width=0.5\textwidth]{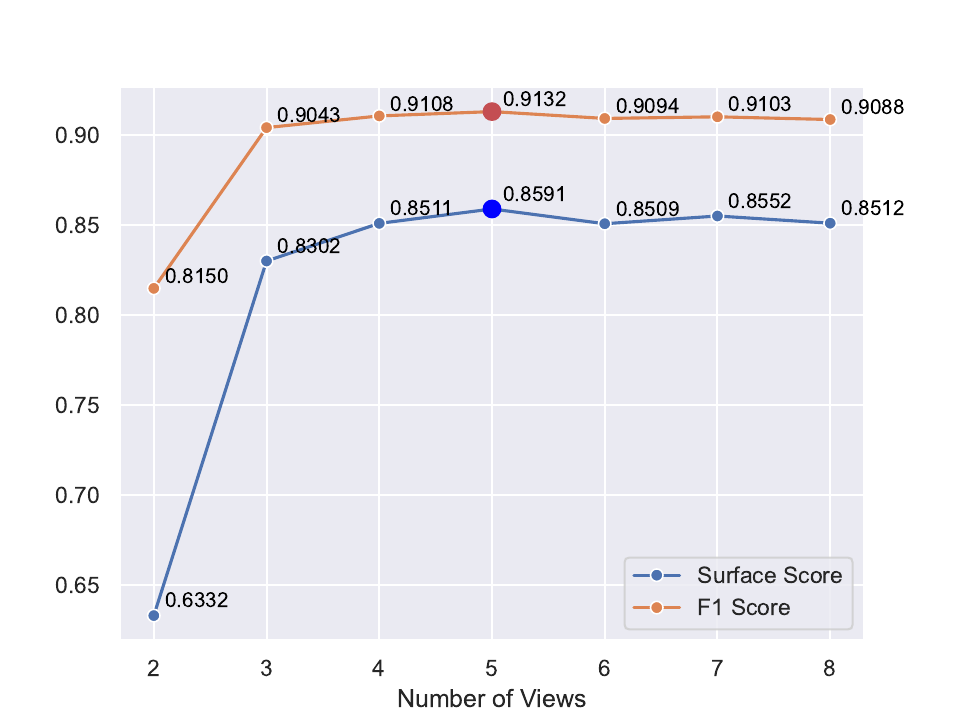}
    \caption{Surface and F1 scores for 3D reconstructions with the \textit{base model} using different input image numbers selected according to Table \ref{tab:number_of_views}.}
    \label{fig:num_of_views}
\end{figure}
Sequentially adding images according to Table \ref{tab:number_of_views}, we obtained the resulting surface and F1 scores, as depicted in Figure \ref{fig:num_of_views}. Both scores followed a similar trend, with the surface score starting lower when using only two images. Both scores plateaued for more than four images, and the best performance was achieved with five images, resulting in an F1 score of 0.9132 and a surface score of 0.8591. Using four instead of five images led to a performance drop of less than $1\%$ for both metrics.

\subsection{View-angle Combination}
\label{sec:results:combination}
Table \ref{tab:view_combinations} displays the experiments conducted on synthetic data to identify the optimal combination of view angles for our 3D reconstruction pipeline. The best surface and F1 scores were obtained by selecting one view from each of the four view classes (AP, lateral, oblique, and miscellaneous). Alternatively, using two AP and two lateral images yielded a comparable F1 score, but the surface score decreased by 4\%. Moreover, employing more AP images than lateral images led to better results than in the reverse scenario.
\begin{table}[htbp]
    \centering
    \begin{tabular}{lcc}
    \hline
    \textbf{View Combinations} & \textbf{Surface $\uparrow$} & \textbf{F1 Score$\uparrow$} \\
    \hline
    1 AP - 1 lateral - 1 OB - 1 MI & \textbf{0.85} & \textbf{0.91} \\
    2 AP - 2 LR & 0.81 & 0.90 \\
    1 AP - 3 LR & 0.70 & 0.85 \\
    3 AP - 1 LR & 0.77 & 0.88 \\
    \hline
    \end{tabular}
        \caption{Performance evaluation of the \textit{base model} across different view angle combinations in the 3D reconstruction pipeline, reported using F1 and surface scores. Best scores are highlighted in bold}
    \label{tab:view_combinations}
\end{table}

\subsection{Sensitiviy to View Angles}
\label{sec:results:sensitivity}
\begin{figure}
    \centering
    \includegraphics[width=0.5\textwidth]{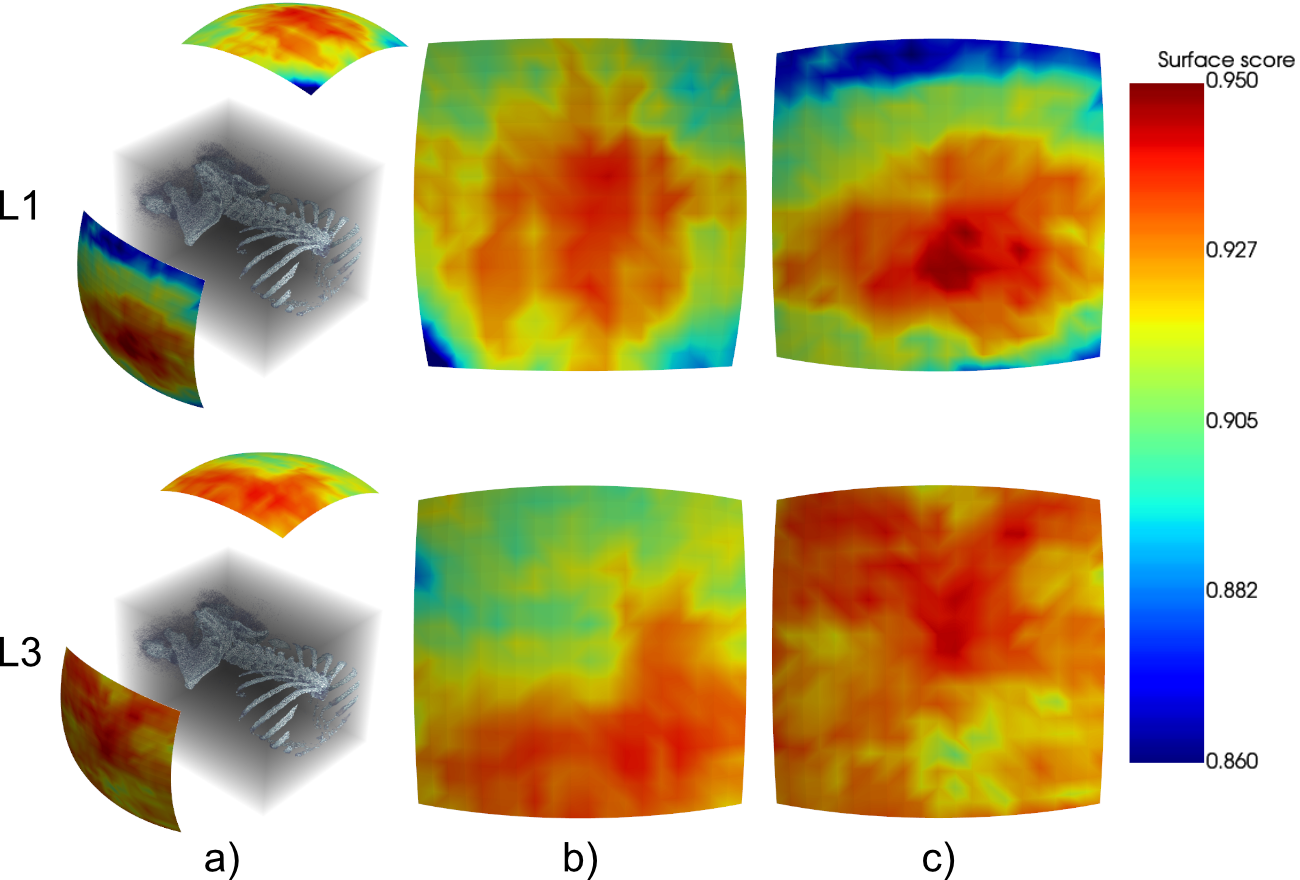}
    \caption{Visualization of surface scores as distance maps on the example of an L1 and L3 vertebra. a) displays the location of the distance maps with respect to the patient. b) represents a heatmap obtained by varying the AP view angle in two directions (moving in the sagittal- and transverse plane),  while keeping the other three images for reconstruction fixed. Similarly, c) shows a heatmap resulting from varying lateral view angles (moving in the coronal- and transverse plane) while keeping the other views constant.}
    \label{fig:L1_L3_distancemap}
\end{figure}

Figure \ref{fig:L1_L3_distancemap} present distance maps depicting the resulting surface scores for varying view angles. The highest surface score achieved was 0.947, surpassing the score reported in \ref{sec:results:reconstruction}. In Figure \ref{fig:L1_L3_distancemap}, both the L1 AP and lateral heatmaps exhibit a radial pattern, with the best scores obtained from the standard clinical AP and lateral views. The scores decay isotropically in all directions. Comparatively, the lateral heatmap in L1 demonstrates more consistent scores for variations within the coronal plane than in the transverse plane. The delta between the highest and lowest scores amounts to $8.6\%$.

L3 depicts slightly different distance maps patterns. The scores also decay in all directions from the highest possible score. With varying AP images, the highest score is, in this particular example, not achieved with a true $90^\circ$ AP image. The lowest achieved score is higher than in the L1 case. The delta between the highest and lowest score amounts $3.8\%$.

\subsection{Challenging Regions}
\label{sec:results:surface_heatmap}
\begin{figure*}
    \centering
    \includegraphics[width=\linewidth]{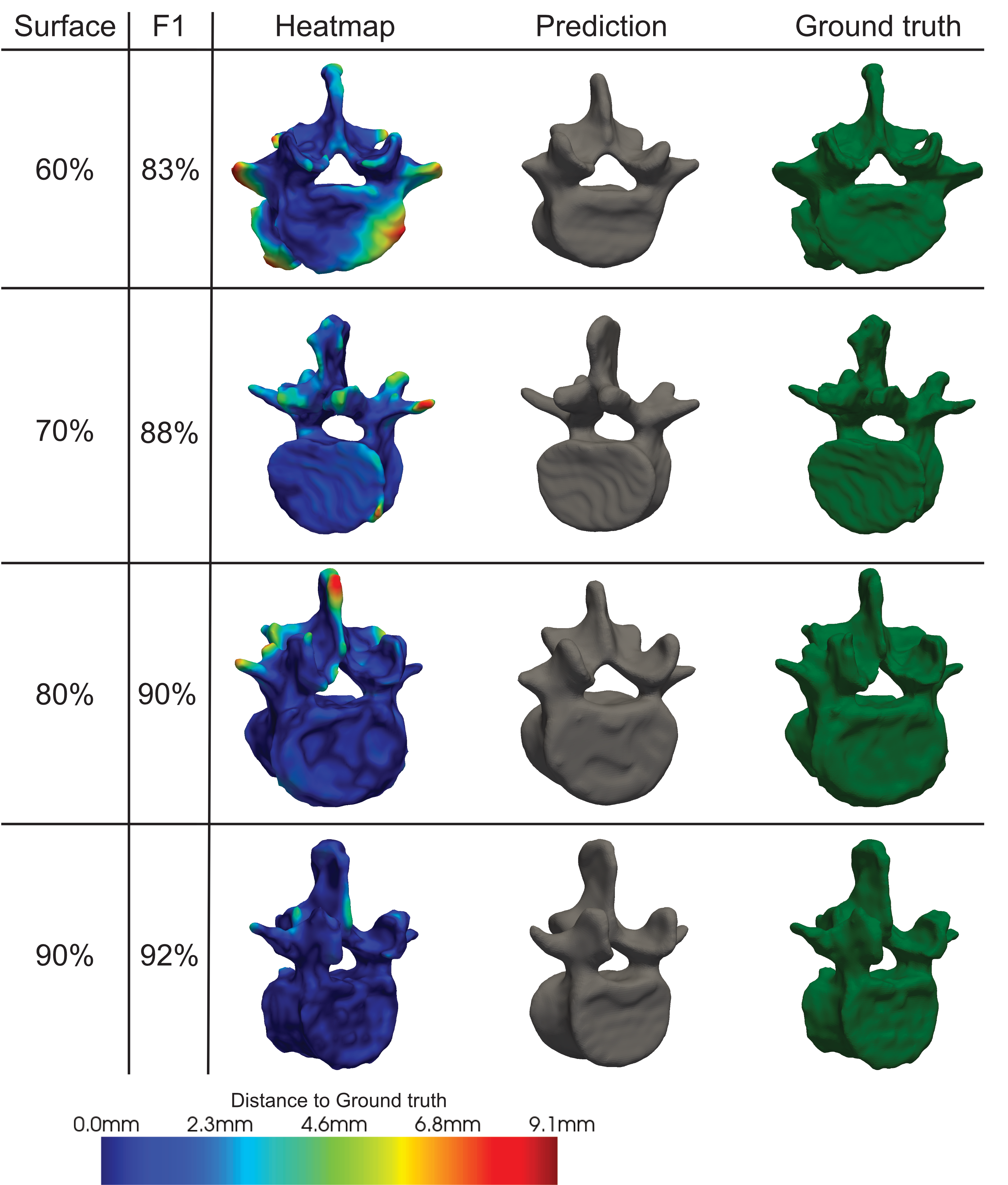}
    \caption{The distance maps depict the spatial deviation in millimeters between the predicted 3D model and the ground truth. Warmer colors, such as red, indicate areas where the prediction deviates from the ground truth, highlighting the challenging regions. On the contrary, cooler colors, such as blue, represent accurately reconstructed areas.}
    \label{fig:60_90}
\end{figure*}

Figure \ref{fig:60_90} presents 3D distance maps illustrating the spatial distances between the predicted models and the ground truth, with a maximum deviation of $9$ mm. 
As highlighted in Section \ref{sec:methods:performance_eval}, while the F1 score is informative for overall vertebrae resemblance, it may not fully capture the intricate and patient-specific details essential for surgical applications. To emphasize this distinction in patient-specificity, we present both the surface and F1 scores along with the distance maps, providing a comprehensive view of the reconstruction quality.
All four reconstructions resemble the 3D morphology of the underlying ground truth, indicated by high F1 scores ranging from $83\%$ to $92\%$.
At the lowest surface score of $60\%$, challenging regions primarily encompass the transverse and spinous processes. Additionally, certain ossifications on the vertebral body are not accurately reconstructed. However, the pedicle regions show accurate reconstruction.

The challenging regions gradually diminish to specific points as the surface score increases. For a surface score of $70\%$, only the tip of one transverse process slightly deviates from the ground truth. In this particular example, with a $80\%$ surface score, a lateral deformity on the left side of the vertebral body is accurately reconstructed.

\subsection{Training Level-specific Models}
\label{sec:results:level_specific}
Separately trained models for each level showed a difference in F1 score only for L1 and L3 (Table \ref{tab:model-comparison}), with a difference of less than $1.1\%$ in both cases. In terms of surface score, the separately trained model performed better only for L1, with a $2.3\%$ performance boost. The \textit{base model} (all-in-one) achieved equal or better scores in all other cases. However, both the separately trained model and the \textit{base model} performed equally on L5 compared to the other four levels.

\begin{table}[htbp]
    \centering
    \begin{tabular}{c|cc|cc}
        \hline
        \multirow{2}{*}{\textbf{Level}} & \multicolumn{2}{c|}{\textbf{Separate Models}} & \multicolumn{2}{c}{\textbf{All-in-One Model}} \\
         & \textbf{Surface $\uparrow$} & \textbf{F1 Score$\uparrow$} & \textbf{Surface $\uparrow$} & \textbf{F1 Score$\uparrow$} \\ 
        \hline
        L1 & \textbf{0.88} & \textbf{0.92} & 0.86 & 0.91 \\ 
        L2 & 0.88 & \textbf{0.92} & \textbf{0.89} & \textbf{0.92} \\ 
        L3 & 0.86 & 0.91 & \textbf{0.87} & \textbf{0.92} \\ 
        L4 & 0.84 & \textbf{0.91} & \textbf{0.85} & \textbf{0.91} \\ 
        L5 & \textbf{0.77} & \textbf{0.89} & \textbf{0.77} & \textbf{0.89} \\ 
        L1-L5 & \textbf{0.85} & \textbf{0.91} & \textbf{0.85} & \textbf{0.91} \\ 
        \hline
    \end{tabular}
    \caption{Comparison of Surface Scores and F1 Scores for Separate Models and All-in-One Model} 
    \label{tab:model-comparison}
\end{table}

\subsection{Paired Dataset Quality Assesment}
\label{sec:results:paired}
We evaluated the quality of our paired data collection approach to highlight the level of spatial alignment between real X-rays and their corresponding DRRs in the generated paired dataset. 
The mean Euclidean distance across all calibration assessments was $1.06 \pm 0.86$ pixels, indicating the average misalignment between user-identified points and the projected 3D points. The median error was $0.86$ pixels. With a pixel spacing of $0.152$ mm, this corresponds to an average error of $0.16 \pm 0.13$ mm and a median error of $0.86$ mm. 

\section{Discussion}
\label{sec:discussion}
This study introduced a novel approach to bridge the domain gap when developing intraoperative deep learning models using synthetic training datasets. We aim to pave the way toward registration-free surgical navigation for the use-case of spine surgery, which presents particular challenges that were individually investigated in the herein study. We showed that real-time and accurate 3D reconstructions are possible for the entire lumbar spine region only using 3 to 4 intraoperative X-rays. To the best of our knowledge, this has not yet been shown in the prior art and can be seen as a significant contribution of the herein work. \textit{Ex-vivo} tests revealed that a seamless transition from \textit{in-silico} training to the real-world application is possible by applying recent domain adaptation algorithms. The main powerhouse that allowed this transition was a paired dataset consisting of real-synthetic image pairs generated thanks to our custom-developed data collection pipeline.

Work, such as the one \cite{tonetti_role_2020}, finds that the benefits of intraoperative 3D imaging outweigh the higher acquisition costs of a CBCT system. The price can be more than three times that of a C-arm for 2D-guided interventions. Among the reasons are the reduced surgical time, higher accuracy, and a reduced amount of revision surgeries. Our work has shown to be capable of creating intraoperative 3D reconstructions with a conventional C-arm.

\subsection{Addressing the Domain Gap}
We investigated two domain adaptation techniques to diminish the domain gap: transfer learning and style transfer methods. The results have shown that the transfer-learned and style transfer approaches perform equally well compared to reconstructions achieved purely using synthetic data. 

The effectiveness of our style-transfer approach demonstrated significant improvements in 3D reconstruction performance on real data compared to previous methods relying solely on synthetic training data. By reducing the domain gap, our approach enhances the practicality and generalizability of 3D reconstruction models, making them suitable for real-world clinical applications. Our style transfer paradigm can be seamlessly integrated into various intraoperative applications where models trained on synthetic data are already available, such as segmentation, reconstruction, and landmark detection, for example, in \cite{gao_synthex_2022, kasten_end--end_2020} and our existing 3D reconstruction network X23D \cite{jecklin_x23dintraoperative_2022}. By applying our style transfer paradigm sequentially, real X-ray inputs are first translated into the synthetic domain and then fed into the pre-trained models.

In the study by \cite{gao_synthex_2022}, our method of emphasizing bones in DRRs through thresholding and integrating intensity values, without considering imaging physics, is termed 'Naive DRR'. They report improved results in landmark detection in hip images using more recent DRR techniques like those in \cite{unberath_deepdrr_2018}, which aim to produce DRRs more closely resembling actual X-rays. This suggests a need for further research to determine if such advanced DRR techniques could enhance the vertebra reconstruction quality of X23D. A potential shortcoming of the target domain we selected for our style transfer approach is its limited versatility for other applications. While emphasizing bones, our target domain is, e.g., less suitable for soft tissue. However, DRRs that more closely match the X-ray domain can be seamlessly integrated into our paired dataset, leveraging the available calibration information of our paired dataset.

\cite{shiode_2d3d_2021} employed a similar style transfer approach using Pix2pix to convert X-rays into the DRR domain. They also relied on the Naive DRR approach, which emphasized the wrist bone for their needs. This underscores the advantages provided by our trained style transfer for lumbar spine images.

\subsection{Performance Evaluation}
Our evaluation compared the performance of three different methods. First, we assessed an exclusively \textit{in-silico} trained \textit{base model}. Second, we examined a \textit{transfer-learned model} using real X-rays, which was built upon the \textit{base model}. Finally, we explored the performance of the \textit{base model} in conjunction with a standalone style transfer model. This style transfer model is designed to convert X-rays to the synthetic domain.

When applied to DRRs, the \textit{base model} exhibited the highest performance across all metrics, indicating the achievable performance without a domain gap. The style transfer approach achieved a $73\%$ surface score. 
Depending on the random combination of four views, results varied. As shown, e.g., in Figure \ref{fig:L1_L3_distancemap}, the network could achieve significantly better scores when provided with good view combinations. 
Section \ref{sec:results:num_of_images} demonstrated that while three images are sufficient to reconstruct a vertebra, adding a fourth image provides a good tradeoff between additional radiation exposure and improved reconstruction quality. Although processing more images can yield further improvements, these are less impactful. High-quality reconstructions using arbitrary perspectives of X-rays are possible with our method; however, having a wider separation between acquisition positions to cover more spatial details is beneficial. Thus, using AP and lateral images in combination with oblique and miscellaneous off-plane X-rays produced the best results, as outlined in Section \ref{sec:results:combination}.
Figure \ref{fig:L1_L3_distancemap} illustrated that deviations from these viewpoints over larger areas only slightly affect reconstruction quality. Deviations on two axes of up to $20^\circ$ did not lower the surface scores in the two examples below $86\%$.
In addition, the quick processing time of $10.3$ms for localization and $81.1$ms for 3D reconstruction allows immediate intraoperative updates. Patient-specific reconstructions, particularly highlighted in Figure \ref{fig:60_90}, underscore the relevance and versatility of our approach for a spectrum of surgical procedures. The distance maps indicate that while deviations are present, they are predominantly in non-critical areas such as the tips of transverse processes. In contrast, critical regions like the pedicles are reconstructed accurately, often with deviations well below 1 mm even at surface scores of $60\%$. This signifies the applicability of our methodology in crucial surgical aspects, such as precise planning for pedicle screw placement.
Further investigation regarding view angle combinations, as shown in Section \ref{sec:results:combination}, could help tackle challenging regions in reconstructions shown in Figure \ref{fig:60_90}.

Despite the inherently more complex topology of vertebrae compared to the ulna, our method outperforms with a marginally better average ASD of $0.97\pm0.20$, compared to the $1.05\pm0.36$ achieved by \cite{shiode_2d3d_2021}. Similar to ours, their approach utilizes style transfer with Pix2Pix and an \textit{in-silico} trained model.

\cite{kasten_end--end_2020}, employing a similar methodology that combines \textit{in-silico} training with cycle GAN style transfer, achieves an F1 score of $94.8\%$ in femur reconstruction from two X-rays. This score is $10.8\%$ higher compared to our approach using the estimated origin. However, the surface scores reflecting patient specificity should also be considered for a comprehensive comparison of the two methods. It is also important to note that data-driven approaches may be more likely to achieve high F1 values for the femur as it approximates its convex hull more closely than a vertebra.

The assessment of the paired-dataset quality showed that the pixel error is slightly greater than the one reported in \cite{zheng_robust_2009}. However, the error fell within the same range due to higher resolution X-rays despite using fewer calibration points. 

\subsection{Influence of Estimated Origin}
Another critical aspect explored in our study was the influence of estimated origins on reconstruction performance. We compared the network's performance using estimated and ground truth origins and observed interesting trends.

When ground-truth origins were used, all approaches exhibited significant performance improvements. The \textit{transfer-learned model} achieved a $3\%$ increase in surface score, while the style transfer approach experienced a remarkable $11\%$ boost. This suggests that both models can enhance their performance significantly when provided with more accurate origin information. Within the estimated origin experiments, the \textit{transfer-learned model} did not perform significantly differently than the style transfer approach using the \textit{base model}. In the ground truth origins experiments, the style transfer approach had a clear and statistically significant advantage over the \textit{transfer-learned model}, with a $7\%$ higher surface score and a $3\%$ higher F1 score. Given the option to train the \textit{base model} using substantially more synthetic training data in the future, the style transfer approach demonstrates enhanced capabilities in terms of accuracy, simplicity, effectiveness for re-training, and further development.

\begin{figure}
    \centering
    \includegraphics[width=\linewidth]{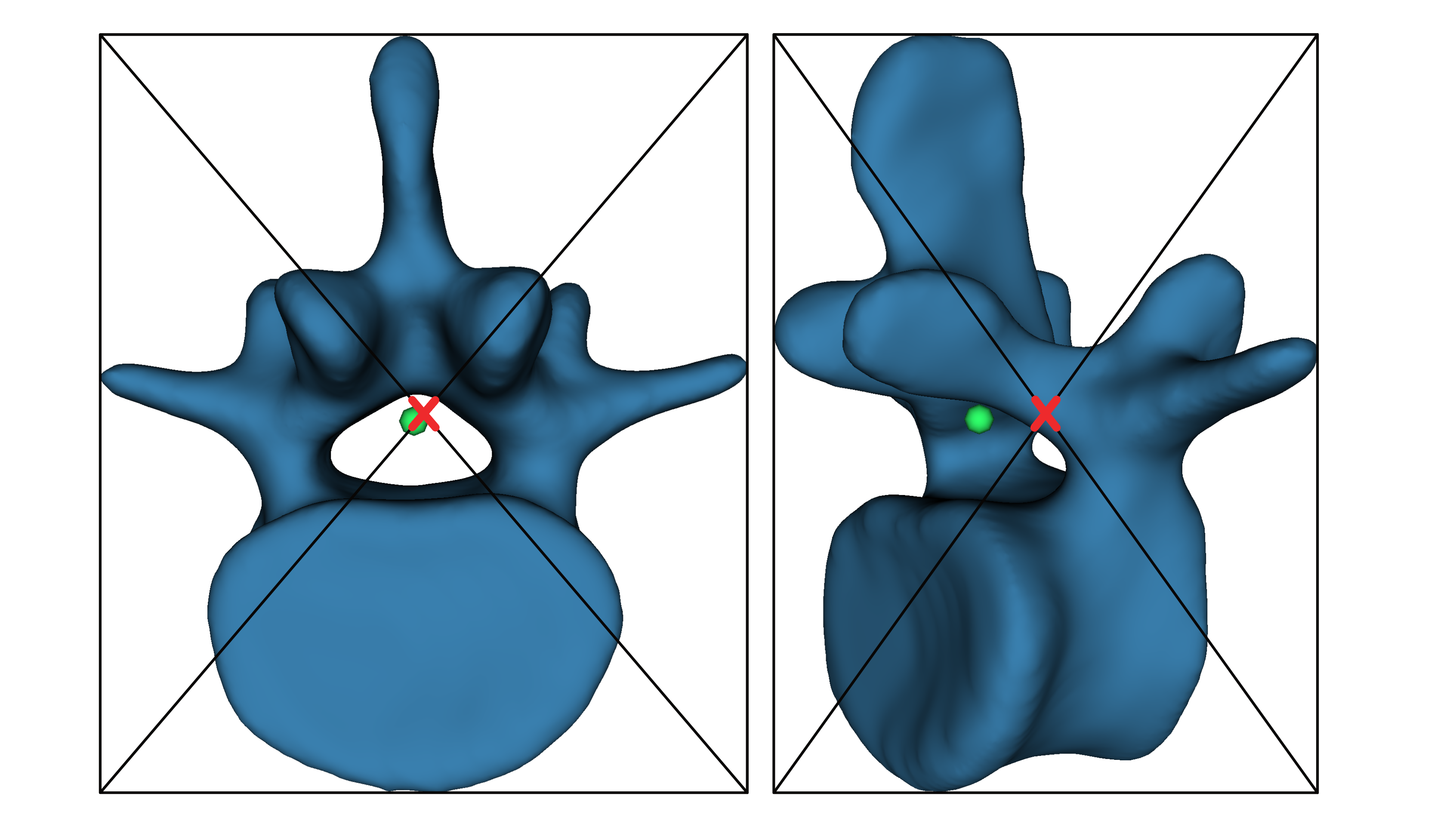}
    \caption{Comparison: Centroid of the Object vs. Center of the 2D Bounding Box.}
    \label{fig:centroid_discussion}
\end{figure}

One problem with our chosen approach for the origin estimation is an error propagation coming from the perspective geometry and a symmetrical assumption. The center of a projected vertebra is not necessarily the center of its 2D bounding box. This principle is visualized in figure \ref{fig:centroid_discussion}. Combined with a narrow baseline between two cameras, this error can propagate and shift the estimated origin significantly. The 3D refiner stage of our network explained in Section \ref{sec:method:network} is used to centered reconstructions. By moving the origin and, therefore, the reconstruction within the reconstruction cube, the 3D refiner can produce 3D reconstructions with varying performance. 

To tackle the influence of the estimated origins, the network can, on the one hand, be improved by conditioning it with augmented data that involves non-centered reconstruction cubes. An additional approach could be a second pass through our network. Our results show that our approach can already produce an accurate reconstruction. A preliminary reconstruction to better estimate the origin could improve the accuracy in a second pass. 

\subsection{Outlook, Limitations and Future Directions}
By reducing the domain gap and achieving accurate 3D reconstructions from real X-rays, our approach opens up new possibilities for intraoperative planning, navigation, and robotic surgery. Surgeons can benefit from high-accuracy intraoperatively generated visualizations, leading to better surgical outcomes.
While our study yields promising results, it is not exempt from limitations. The effectiveness of our approach is contingent on the precision of the localization step, which can be affected by factors like patient positioning and imaging conditions. Ongoing research is imperative to tackle these challenges and enhance the resilience of our approach in real clinical scenarios. 

To eliminate the need for a calibration phantom used during inference as described in section \ref{sec:method:network}, future research might investigate the application of structure-from-motion (SFM) principles or recent techniques like those proposed in \cite{yen-chen_inerf_2021} for determining relative poses between X-rays.

The current YOLO-based localization method has proven effective, with satisfactory results and the option for surgeons to correct false detections manually. However, future work will explore more sophisticated methods that integrate data from multiple viewpoints to enhance the robustness and accuracy of vertebra localization. For example, techniques from multi-view vertebra localization studies, such as the one by \cite{wu_multi-view_2023}, show promise for improving label accuracy. Additionally, by training future localization networks in the DRR domain, we can leverage larger datasets for more comprehensive training. Our style transfer network enables us to apply these advanced localization techniques, ensuring smoother and more reliable vertebra detection in diverse clinical settings.

Additionally, there is room to enhance our current method of origin estimation. Improvements in this area could simplify the task for the 3D refiner stage, thereby increasing the overall accuracy of the 3D reconstruction.

Looking ahead, we aim to integrate our 3D reconstruction pipeline into CAS systems and undertake comprehensive clinical trials. These evaluations are required to determine the pipeline's impact on surgical accuracy and outcomes.

One notable limitation when transferring our approach to a CAS system is the potential impact of breathing motion on the accuracy of 3D reconstructions. Breathing-induced movement can vary significantly across different regions of the spine. The thoracic region, being closer to the lungs, is more susceptible to movement, whereas the lumbar region, which is the primary focus of our study, is generally less affected.
\cite{glossop_assessment_1997} reported a movement of up to 1.3 mm in the L3 and L4 vertebrae during spine surgery, with a mean deviation of 1.59 mm in the cervical region and 1.11 mm in the thoracic region. \cite{winklhofer_spinal_2014} differentiated between normal and forced breathing, finding motion of 0.34 mm and 1.4 mm, respectively, in the lumbar region during spine surgery. These findings suggest that while there is some deviation due to breathing, the lumbar spine experiences relatively moderate movement. However, this deviation is relatively larger than our reported average surface distance, indicating the need for further investigation.
Such analyses are considered out of the scope of the herein study, which is primarily \textit{ex-vivo} based. In the future phases of this project and through comprehensive \textit{in-vivo} testing, amongst other factors, we will focus on assessing the severity of breathing-induced deviations and investigating potential countermeasures such as a patient tracking systems, to compensate for this motion and enhance the accuracy and reliability of our method in clinical settings.

Although our approach is generally unaffected by artifacts present in the input X-rays, provided they do not occlude the vertebrae to be reconstructed, there are clinical scenarios where metal artifacts such as screws and k-wires will be present and must be considered. Additionally, the current approach has only been trained on a wide range of \textit{ex-vivo} specimens without systematic evaluation of the extent of their pathologies. In our future work, we will enhance the training data by incorporating metal artifacts and systematically test our method across various pathological conditions. This will improve the robustness and broaden the clinical applicability of our method.

\section*{Data Availability}
The code presented in this study is available on request from the corresponding author. The code is not publicly available due to a pending patent application. The base for the described synthetic dataset is available here: \url{https://github.com/MIRACLE-Center/CTSpine1K} (accessed on 25 January 2024).

\section*{Declaration of Competing Interest}
The authors declare the following financial interests/personal relationships which may be considered as potential competing interests:
Sascha Jecklin reports financial support was provided by the Monique Dornonville de la Cour Foundation. Sascha Jecklin reports financial support was provided by an internal Balgrist fund. Mazda Farshad reports a relationship with X23D AG that includes: equity or stocks. Hooman Esfandiari and Philipp Fürnstahl report a relationship with X23D AG that includes: board membership and equity or stocks. Hooman Esfandiari, Philipp Fürnstahl and Mazda Farshad have a patent \#WO2023156608A1 pending to University of Zurich. Hooman Esfandiari, Philipp Fürnstahl, Mazda Farshad, and Sascha Jecklin have a  patent "A computer-implemented method, device, system and computer program product for processing anatomic imaging data" pending to University of Zurich.

\section*{CRediT Authorship Contribution Statement}
\textbf{Sascha Jecklin:} Conceptualization, Methodology, Software, Validation, Formal analysis, Investigation, Data Curation, Writing - Original Draft, Visualization. \textbf{Youyang Shen:} Methodology, Software, Validation. \textbf{Amandine Gout:} Methodology, Software, Validation. \textbf{Daniel Suter:} Investigation. \textbf{Lilian Calvet:} Software, Writing - Review \& Editing. \textbf{Lukas Zingg:} Investigation. \textbf{Jennifer Straub:} Investigation. \textbf{Nicola Alessandro Cavalcanti:} Investigation. \textbf{Mazda Farshad}: Resources. \textbf{Philipp Fürnstahl:} Conceptualization, Resources, Writing - Review \& Editing, Supervision, Funding acquisition. \textbf{Hooman Esfandiari:} Conceptualization, Methodology, Software, Writing - Original Draft, Writing - Review \& Editing, Supervision, Project administration.

\section*{Acknowledgements}
This work has been supported by the OR-X - a Swiss national research infrastructure for translational surgery and associated funding by the University Hospital Balgrist. We like to thank Tanja Walther for her support during the data capture.

\section*{Ethical Approval}
The study was conducted according to the guidelines of the Declaration of Helsinki, and approved by the local ethical committee (KEK Zurich BASEC No. 2021-01083)

\section*{Declaration of Generative AI and AI-assisted Technologies in the Writing Process}
During the preparation of this work the author(s) used Grammarly and ChatGPT in order to improve the text quality. After using this tool/service, the author(s) reviewed and edited the content as needed and take(s) full responsibility for the content of the publication.

\bibliographystyle{model2-names.bst}\biboptions{authoryear}
\bibliography{main}

\end{document}